\documentclass[sn-mathphys,Numbered]{sn-jnl-arxiv}%

\usepackage{natbib}
\usepackage{graphicx}%
\usepackage{subcaption}%
\usepackage{multirow}%
\usepackage{amsmath,amssymb,amsfonts}%
\usepackage{amsthm}%
\usepackage{mathrsfs}%
\usepackage{xcolor}%
\usepackage{textcomp}%
\usepackage{manyfoot}%
\usepackage{booktabs}%
\usepackage{algorithm}%
\usepackage{algorithmicx}%
\usepackage{algpseudocode}%
\usepackage{listings}%
\usepackage{pdflscape}%
\usepackage{url}%
\usepackage{fancyvrb,fvextra}
\usepackage{array}
\usepackage{longtable}

\begin{document}

\title[AI-Assisted Hypothesis Generation for Hospital QI]{AI-Assisted Hypothesis Generation for Hospital Quality Improvement}

\author*[1,2]{\fnm{Patrick} \sur{Vossler}}\email{patrick.vossler@ucsf.edu}
\equalcont{These authors contributed equally as co-first authors.}
\author*[1,2]{\fnm{Jean} \sur{Feng}}\email{jean.feng@ucsf.edu}
\equalcont{These authors contributed equally as co-first authors.}
\author[1]{\fnm{Venkatesh} \sur{Sivaraman}}
\author[1]{\fnm{Robert} \sur{Gallo}}
\author[1,2]{\fnm{Hemal} \sur{Kanzaria}}
\author[2]{\fnm{Dana} \sur{Freiser}}
\author[2]{\fnm{Christopher} \sur{Ross}}
\author[1,2]{\fnm{Amy} \sur{Ou}}
\author[1,2]{\fnm{James D.} \sur{Marks}}
\author[2]{\fnm{Susan} \sur{Ehrlich}}
\author[1,2]{\fnm{Christopher} \sur{Peabody}}
\author[1,2]{\fnm{Lucas} \sur{Zier}}

\affil[1]{\orgdiv{University of California, San Francisco}, San Francisco, CA, USA}
\affil[2]{\orgdiv{Zuckerberg San Francisco General Hospital}, San Francisco, CA, USA\vspace{4pt}}

\abstract{
Hospital quality improvement (QI) programs generally proceed in two stages: (1) hypothesis generation, in which candidate factors that may improve care quality and efficiency are identified, and (2) confirmatory analyses, in which specific QI initiatives targeting these factors are implemented and evaluated.
While most research focuses on the confirmatory stage, the first stage of hypothesis generation and exploratory data analysis is equally important as it shapes where teams prioritize their effort for the highest impact.
Currently, hypothesis generation for QI depends on expert-driven, qualitative methods (e.g., tools from Lean Healthcare) that can be anecdotal, hard to reproduce, time-consuming, and resource-intensive.
LLMs have the potential to substantially accelerate these processes and strengthen their evidentiary basis by analyzing clinical notes at scale, but existing AI tools can only scale up tasks that are already well-defined, whereas hypothesis generation is by definition a ``fuzzy,'' exploratory, and iterative process.
To learn how to automate QI hypothesis generation, we instead propose treating not only LLM instructions but also the problem specification and validation process as tunable natural-language (hyper)parameters.
Through a procedure we term ``Human-AI Spec-Solution Co-Optimization,'' we iteratively elicit clinical experts' definitions and judgments, while aligning the AI pipeline with their intended criteria.
We retrospectively applied this framework at an urban safety-net hospital for two operational metrics: length of stay and 30-day unplanned readmissions.
The final AI pipelines first screen for potential factors in individual patient charts and then filter for clinically relevant factors through self-rating, where AI self-ratings achieved over 70\% concordance with human experts (within one Likert point) on held-out encounters.
When scaled up to understand hospital-wide factors, the pipelines identified a superset of the candidate factors found in prior, primarily manual analyses.
Furthermore, by providing auditable reasoning traces and quantifying the prevalence of candidate factors across analyzed encounters, the pipelines facilitated a more quantitative and readily reproducible approach for choosing which factors to prioritize for prospective QI intervention design, implementation, and testing.
}


\maketitle

\section{Introduction}

Quality Improvement (QI) programs play a critical role in optimizing hospital efficiency and patient outcomes by identifying and mitigating gaps in healthcare delivery \citep{D-Andreamatteo2015-zg, Graban2018-ac, Mercer2019-qj, Chase2020-vy, Schwartz2024-sk}.
Hospital QI typically begins with planning, where teams explore, identify and prioritize candidate factors, then moves on to implementation of specific interventions and evaluation through confirmatory analyses.
While the latter phase has garnered much of the attention in the clinical literature, planning is essential to successful QI efforts: poor planning is one of the most frequently cited causes for QI programs ultimately failing to achieve their goals~\citep{Shojania2005-px, Mauck2011-id, Tallentire2019-qp,Longenecker2014-wh}.
Therefore, widely-used QI methodologies such as Lean Healthcare require first building a contextually-grounded understanding of the current state of a system, through an extensive hypothesis generation process~\citep{Womack2005-nw}.
However, hypothesis generation for QI is complex, underspecified (or ``fuzzy''), and currently limited in its scalability, transparency, and reproducibility \citep{Po2019-gw, Rundall2021-nb, Marsilio2022-hp}.
QI teams typically adopt a combination of semi-structured qualitative tools including fishbone diagrams, chart reviews, staff interviews, value stream mapping, and gemba visits \citep{Bush2007-fh, Catalyst2018-ho, Steverson2023-hu, Quiroga2025-uu}.
Each technique captures a different view of hospital processes, but also requires time-consuming iterations and expert commitment.
As an illustration, consider a recent effort at the urban safety-net Zuckerberg San Francisco General Hospital (ZSFG) to identify major drivers for prolonged length of stay (LOS).
This 6-month effort surfaced numerous questions, in an often unexpected and inconvenient order: ``What factors are considered modifiable?'', ``Which experts should be interviewed and who is available?'', ``Which patient charts should we review and how many, given the team's limited resources?'', and ``What is the prototypical patient journey and which elements should be included in a Value Stream Map?''.
These questions were primarily answered based on expert judgment, where many decisions were inevitably implicit, ill-specified, and subject to cognitive biases.
Furthermore, this process would be difficult to retrace if applied at a different hospital or even repeated at the same hospital.

These limitations suggest a need for more systematic, data-driven approaches for conducting QI hypothesis generation.
\textit{In particular, Artificial Intelligence (AI), and more specifically Large Language Models (LLMs), has the potential to transform the hypothesis-generation phase of QI by streamlining some of the most time- and labor-intensive tasks to be highly efficient, generalizable, and reproducible.}
In particular, recent findings show that LLMs possess strong clinical reasoning abilities and can extract clinical concepts with high accuracy from lengthy medical documents \citep{Agrawal2022-me, Guevara2024-ul, Li2026-sg}, creating an opportunity to scale chart review beyond what is feasible manually.
In addition, AI pipelines can produce auditable artifacts, such as reasoning traces and links to supporting evidence in patient charts, that make the analytical process more inspectable and transparent than many manual reviews.
Finally, AI pipelines are also fully documented in the form of prompts, code, and hyperparameters, making such analyses considerably easier to reproduce than conventional hypothesis generation methods. Rerunning the analysis is straightforward as new operational questions arise.
However, typical methods for aligning AI pipelines, such as manual or automated prompt tuning \citep{Khattab2023-yj, Yang2024-dt} and fine-tuning \citep{Ding2023-dq}, are insufficient for scaling up QI hypothesis generation.
These methods typically assume the information extraction task is fixed and well-specified, with a fixed set of desired elements, clear definitions, or a dataset with gold-standard labels.
QI hypothesis generation satisfies none of these requirements, because inherent to the exploratory process is that experts are exploring and learning the exact set of problem specifications as they go.
This makes validation particularly difficult, since the goal is to design an AI pipeline that matches human \textit{latent} intent.
This challenge, recently termed the ``gulf of envisionment'' \citep{Bubenko2002-xs, Subramonyam2024-at}, is typically addressed through an iterative process in which humans provide feedback to the AI agent.
Still, prior works focus on ``smaller'' gulfs with only one or two underspecified components that can be addressed through bespoke solutions \citep{Kothari2026-zo, Feng2026-ip}, whereas nearly every component is underspecified in QI hypothesis generation.

\begin{figure}[ht!]
\centering
\includegraphics[width=0.8\linewidth]{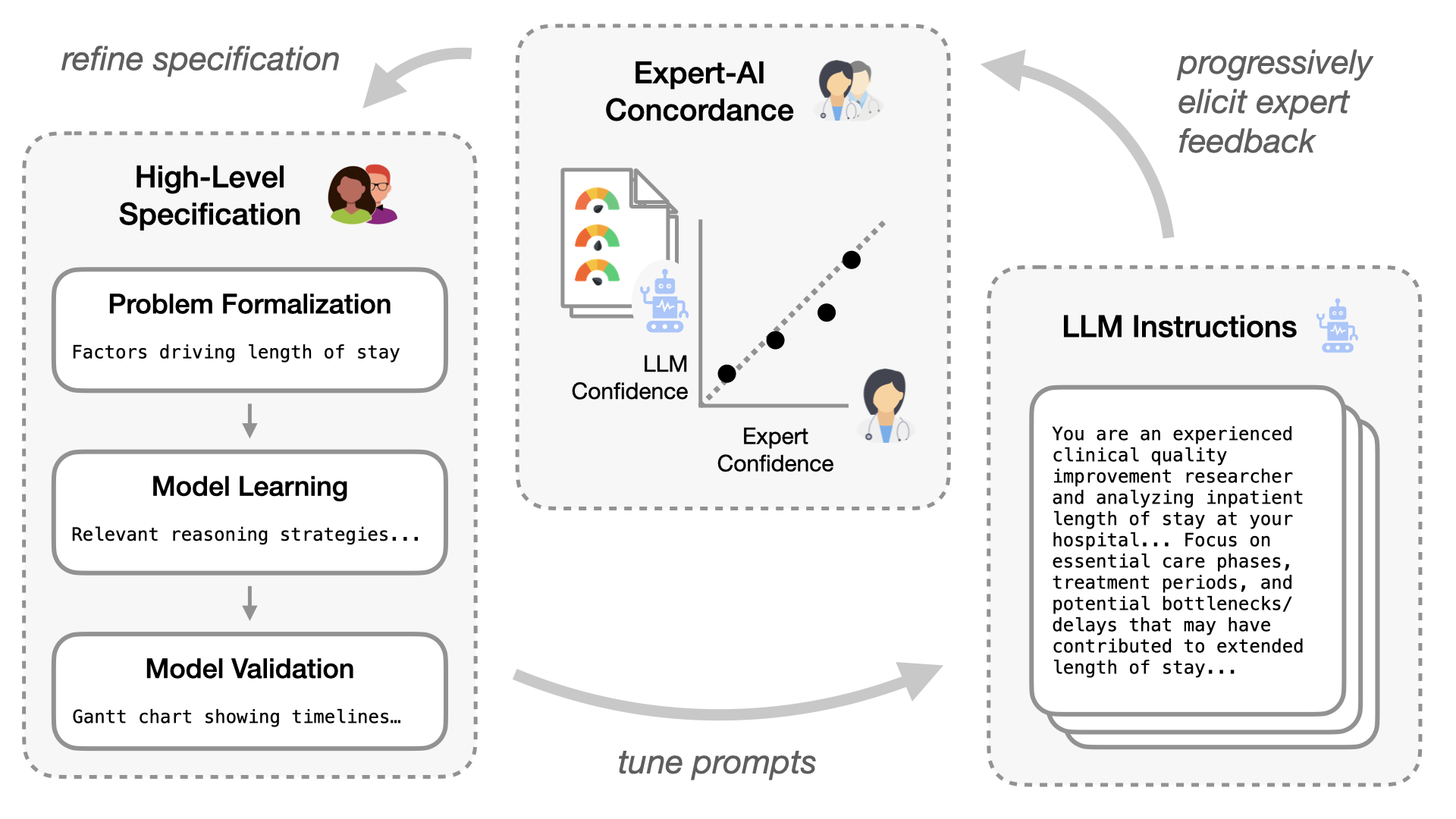}
\caption{Scaling up QI hypothesis generation using AI requires acknowledging that it is inherently complex, exploratory, and iterative, with experts learning as they go. To design AI pipelines that are faithful to this exploratory process, we need to tune not only LLM prompts (right box) but also high-level specifications regarding the exact problem and approach, structured as steps of the AI/ML development process (left box).
The high-level specifications and LLM prompts are then iteratively tuned with input from both the QI team and AI agent to optimize the concordance between both parties (middle box), in a process of ``Human-AI Spec-Solution Co-Optimization.''}
\label{fig:joint_optimization}
\end{figure}

QI hypothesis generation demands a more general approach that can simultaneously accommodate free-form exploration and ensure scientific rigor.
We propose treating not just LLM prompts but also the high-level specifications about the problem, pipeline, and validation approach as parameters to be optimized.
To structure this optimization, we map the specifications to three steps of classical AI/ML development: (1) problem formalization, (2) model learning, and (3) model validation.
Under this mapping, the specifications become natural-language-valued hyperparameters, and the overall problem reduces to one of hyperparameter optimization.
We then introduce an efficient human-in-the-loop optimization procedure, which we call \textit{Human-AI Spec-Solution Co-Optimization}, in which domain experts and the AI agent collaboratively refine both the high-level specifications and the LLM prompts to improve human-AI concordance in the outputs (Figure~\ref{fig:joint_optimization}). %
Expert review of each specification helps ensure that refinements remain clinically meaningful.
The resulting specifications can then be presented to stakeholders alongside the discovered factors, giving them full visibility into the choices that shaped the analysis.
Finally, a held-out set is used to evaluate the tuned pipeline to ensure the final results are unbiased.
We evaluate this framework at ZSFG to develop AI pipelines for two hospital metrics: LOS and 30-day unplanned readmissions.
On held-out data, the system achieved 76.7\% and 71.0\% concordance with expert annotations within one Likert point for the two metrics, dramatically increased patient coverage relative to prior Lean analyses that were primarily manual, and produced auditable reasoning traces.
The pipelines recovered all categories previously identified through prior Lean work at the hospital and surfaced additional candidate factors; together, these findings are being used to design prospective QI evaluations at the hospital.

\section{Results}
\label{sec:results}

Our goal is to scale up QI hypothesis generation so that the learned AI pipeline surfaces candidate factors that experts are most likely to suggest based on the data.
The central difficulty is that hypothesis generation is fundamentally a formative process in which high-level goals are iteratively distilled into actionable steps in an often, and necessarily, ill-formed, exploratory, and implicit manner.
Consequently, existing approaches like manual or automatic prompt tuning methods \citep{Khattab2023-yj, Ding2023-dq, Yang2024-dt} are inapplicable, because they assume the task is well-defined and a gold-standard set of factors exists.

To enable simultaneous exploration and learning of the QI hypothesis generation process, our key insight is to treat LLM prompts as natural-language-valued parameters to be learned while defining the high-level specifications of the problem and validation process as additional natural-language-valued hyperparameters to be explored and refined.
Furthermore, we provide a structured framework by mapping QI hypothesis generation onto the three classical stages of AI/ML development---\texttt{Problem Formalization}, \texttt{Model Learning}, and \texttt{Model Validation}---each broken into sub-components such as the analysis objective, patient population, pipeline inputs and outputs, LLM family/version, and how outputs are validated (Table~\ref{tab:generalized_pipeline}).
Consequently, the task of aligning an AI pipeline for QI hypothesis generation can be viewed as optimization of both natural-language hyperparameters and parameters, the analog of numerical hyperparameters and parameters in classical AI/ML development.

\begin{figure}[ht!]
\centering
\includegraphics[width=\linewidth]{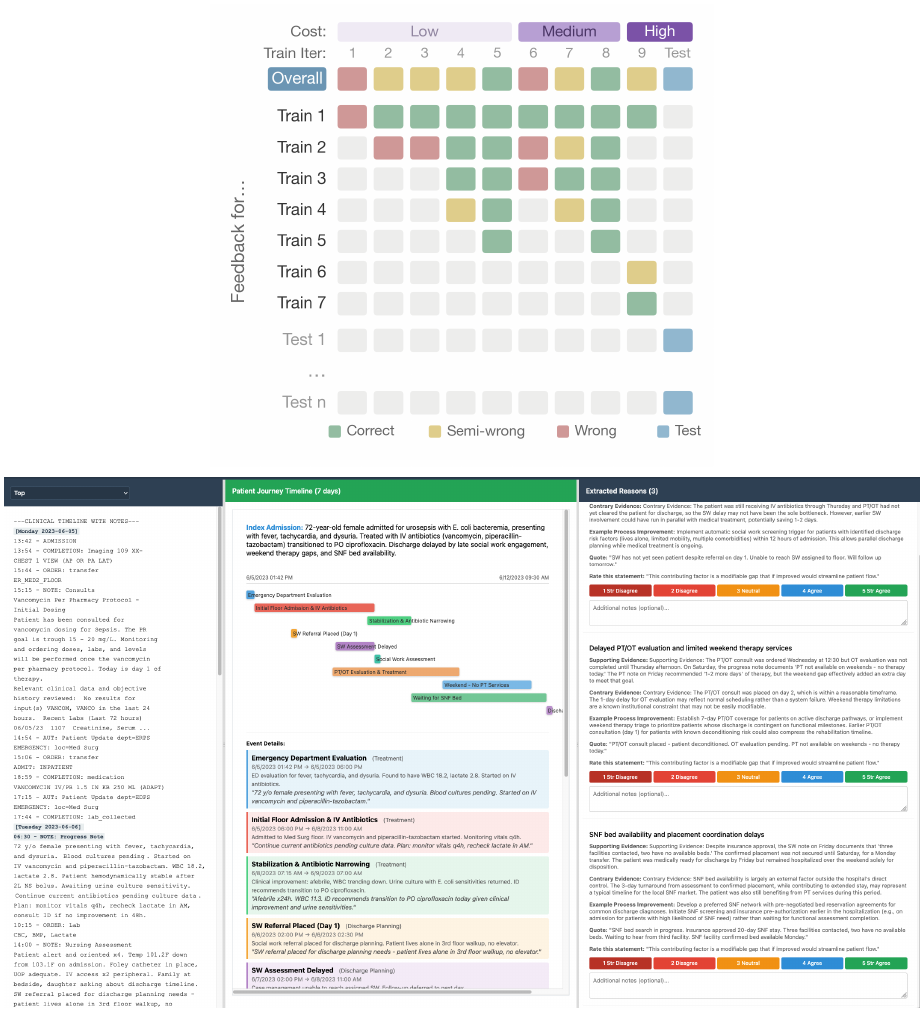}
\caption{
\textbf{Top}:  Example of an online implementation of Human-AI Spec-Solution Co-Optimization, which is more time- and resource-efficient than collecting all annotations upfront.
Rows are patient cases used during training (Train~1--7) and held out for final evaluation (Test~1--n); columns are successive training iterations in which the team refined specifications and prompts.
Each cell is colored by whether the AI pipeline's output on that case was judged correct, semi-wrong or wrong at each iteration.
The two nested loops govern the process.
The outer loop (top ``Overall'' row, shaded by Cost) progressively escalated the level of human expertise used to generate labels, from a single data scientist (low cost and quality) to a clinician (medium cost and quality) to a full group review (high cost and quality).
The inner loop (Train rows) increased the amount of data labeled at each expertise level. Test cases are scored only in the final column.
\textbf{Bottom}: To facilitate validation collaboration between the QI team and AI agent as part of Human-AI Spec-Solution Co-Optimization, a user interface was designed that surfaces not just the raw data (left panel) but also evidence extracted by the AI along its full reasoning path, at multiple levels of abstraction. This includes a Gantt chart representation of the patient chart (middle panel) as well as potential modifiable gaps (right panel), with links mapping back to quotes from the raw data.
}
\label{fig:workflow_ui}
\end{figure}

We optimize these specifications and LLM prompts through a procedure we refer to as \textit{Human-AI Spec-Solution Co-Optimization} (Figure~\ref{fig:joint_optimization}).
Each round, the team executes and evaluates the AI pipeline per the current specifications, refines the over-arching problem specifications based on the elicited feedback, and tunes the prompts per the updated specifications.
Each iteration is collaborative between the QI team and the AI agent, where the latter often helps accelerate validation and refinement.
To maximize the efficiency of the optimization process, we leverage an \textit{online}, adaptive learning approach where we increase the budget of the annotation in each iteration, both in terms of the number of observations annotated and the quality of the feedback/labels from higher-level experts (Figure~\ref{fig:workflow_ui} top).
We also developed a PHI-compliant interface that accelerates the review of AI outputs by presenting the patient chart at multiple levels of abstraction, including the raw data, a map of the patient journey, and summaries of both supporting and contrary evidence (Figure~\ref{fig:workflow_ui} bottom).
The Supplementary Information details each round of Human-AI Spec-Solution Co-Optimization that took place during our case studies.

\begin{table}[!ht]
\centering
\small
\caption{
Mapping the task of QI hypothesis generation to steps of the AI/ML development process.
}
\label{tab:generalized_pipeline}
\label{tab:gulf-mapping}
\begin{tabular}{p{2.2cm}p{6cm}p{6cm}}
\toprule
\textbf{Step} & \textbf{QI Mapping} & \textbf{Example of underspecification} \\
\midrule
\multicolumn{3}{l}{\textit{1. Problem Formalization}} \\
\midrule
Objective & The QI team's goal of the QI factor analysis & Is the goal to find factors that improve hospital flow or overall LOS? \\
\addlinespace
Population & The patients or cases to include in the analysis & How exactly are inclusion/exclusion criteria defined? \\
\addlinespace
Label definition & The definition of a ``correct'' label & Do we generate separate scores for whether a factor is modifiable and causal? \\
\midrule
\multicolumn{3}{l}{\textit{2. Model Learning}} \\
\midrule
Estimator inputs & Input data for the AI pipeline & Which data and notes to pull? \\
\addlinespace
Estimator output & Final and intermediate outputs of the AI pipeline & Should the AI pipeline output confidence scores plus reasoning traces? \\
\addlinespace
Model family selection & The choice of LLM & Which model is the team allowed to use? \\
\addlinespace
Prompt tuning & The prompt text and structure, including multi-step and multi-turn configurations & Which ICL examples to include? How much context can one squeeze? \\
\midrule
\multicolumn{3}{l}{\textit{3. Model Validation}} \\
\midrule
What gets validated & The outputs to validate & Should the annotator validate both extracted reasons and supporting evidence? \\
\addlinespace
How is output validated & Who validates the model, how, and how reliable are the resulting labels? & Expert agreement annotations on LLM-extracted reasons. \\
\bottomrule
\end{tabular}
\end{table}

We scaled up QI hypothesis generation at ZSFG for two hospital metrics: LOS and 30-day unplanned readmission rate.
The team responsible for developing the AI pipelines consisted of one data scientist and six healthcare providers, including physicians, nurses, and QI leadership with Lean expertise.
For each case study, we describe the learned pipelines' concordance with expert ratings on held-out observations, how the iterative learning process unfolded, and the candidate factors identified when the tuned pipeline was applied to the larger held-out cohort.
As the goal of the AI pipeline is to support and scale human-driven hypothesis generation, the primary performance metric was \textit{concordance} between the AI pipeline and annotators when rating candidate QI factors, measured as the rate at which the score implied by the AI's confidence matches the score a human reviewer assigns to the same factor, either exactly or within one point on a five-point Likert scale.
Because early-stage QI hypothesis generation lacks independent ground-truth labels, concordance should be interpreted as a workflow-reliability measure that assesses whether the pipeline surfaces plausible, clinically interpretable candidate factors aligned with expert judgment, not whether those factors are causal or whether interventions targeting them improve care processes or outcomes.
Furthermore, the concordance metric should be interpreted based on the validation workflow used.
As part of the Human-AI Spec-Solution Co-Optimization process, teams explored various validation workflows, where they ultimately settled on an AI-supported approach that optimizes the tradeoff between practical feasibility and the risk of anchoring bias.
These decisions are transparently documented as part of the co-optimization process to ensure appropriate interpretation of final validation results.
Finally, as a secondary indicator of generated hypothesis quality, we qualitatively compared themes identified from the AI pipelines against prior Lean analyses conducted at ZSFG that primarily relied on manual expert review.

\subsection{Case Study 1: Length of Stay}

To characterize modifiable contributing factors for extended LOS, the QI team was tasked with analyzing adult patients across the top five DRGs in the hospital (Sepsis, Skin and Soft Tissue Infection, Ischemic Stroke, Blunt Head Injury, Drug Poisoning), focusing on patients whose LOS was between 4 and 20 days (excluding routine short stays and medically complex outliers).
The high-level problem specifications and their evolution over the Human-AI Spec-Solution Co-Optimization process are summarized in Figure~\ref{fig:los_heatmap}, which shows how specifications for every step of the AI/ML development process were found to need refinement.

\begin{figure}[ht!]
\centering
\includegraphics[width=\linewidth]{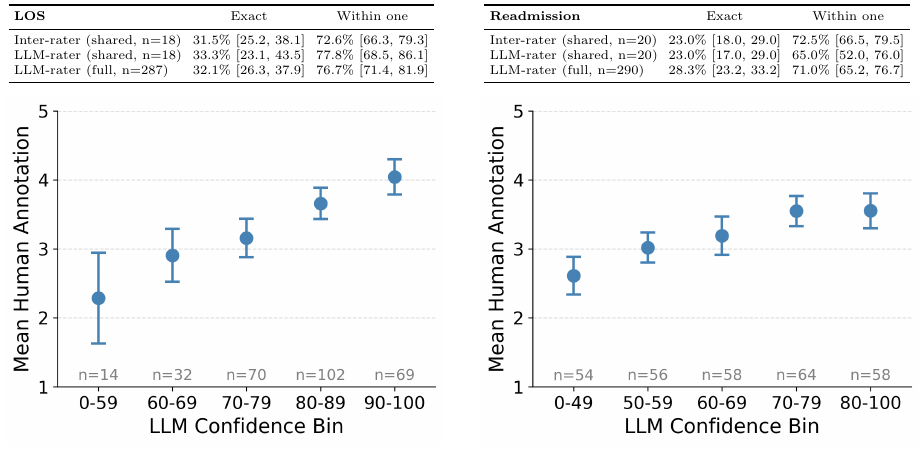}
\caption{AI pipeline evaluation and calibration results for LOS (left) and 30-day unplanned readmission (right).
Each panel shows the agreement table (top) and the calibration curve (bottom) for one case study.
Tables: agreement on the shared subset for factors rated by all reviewers and the full set of factors in the held-out evaluation set.
For shared rows, n is the number of factors rated by all reviewers (six for LOS, five for readmission), with agreement pooled over all rater pairs (inter-rater) or LLM-rater pairs (LLM-rater); for full rows, n is the number of expert ratings.
Brackets are 95\% bootstrap confidence intervals.
Curves: mean expert rating by LLM confidence bin, with 95\% intervals and per-bin sample sizes.
}
\label{fig:pipeline_eval}
\end{figure}

On a held-out dataset annotated by human experts, the final pipeline was first evaluated on 18 factors rated by all six reviewers to compare LLM-rater agreement with inter-rater agreement.
Due to the subjectivity of QI factor rating, exact agreement for both LLM-rater and inter-rater was low (33.3\% and 31.5\%, respectively).
Agreement within one point on the five-point Likert scale, the second concordance measure defined above, better captures meaningful alignment between raters.
The within-one-point agreement rates were 77.8\% for LLM-rater versus 72.6\% for inter-rater agreement, and the difference was not statistically significant (95\% CI for the difference $-6.9$ to $16.1$).

\begin{landscape}
\begin{figure}[ht!]
\centering
\includegraphics[width=\linewidth]{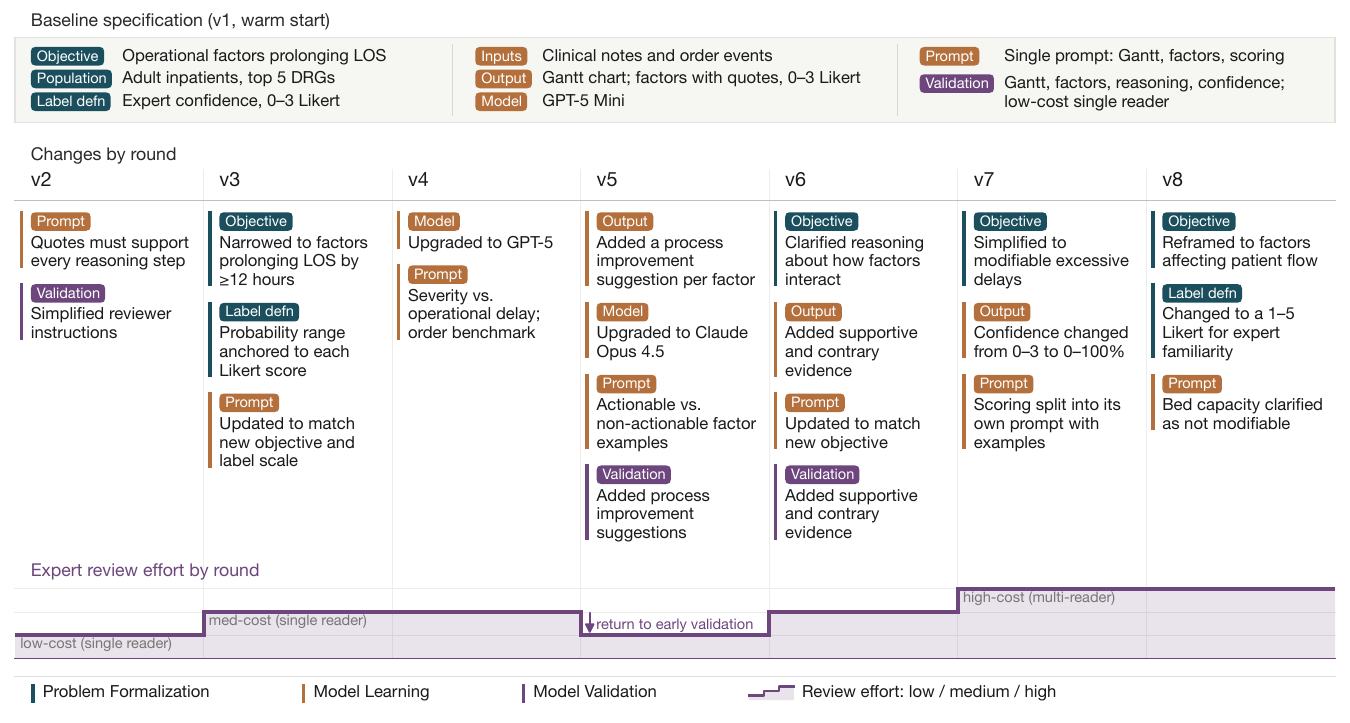}
\caption{Pipeline refinements across LOS optimization rounds.
The top panel shows the initial (v1) specification; each column below lists the components changed in one round, colored by AI/ML development stage (Table~\ref{tab:generalized_pipeline}); components not listed were carried forward unchanged.
The bottom panel tracks expert review effort by round; the return to early validation at v5 is discussed under Case Study 1 in Results.
Full per-round specifications and the feedback behind each change appear in the Supplementary Information.
}
\label{fig:los_heatmap}
\end{figure}
\end{landscape}

The AI pipeline was then evaluated in a larger annotated dataset of 52 assigned patients and 287 total annotations.
Agreement within one point was 76.7\% (95\% CI 71.4--81.9\%); additional agreement statistics and sensitivity analyses are in the Supplementary Information.
The calibration curve (Figure~\ref{fig:pipeline_eval} left) confirmed that higher LLM confidence corresponded to higher expert ratings.
LLM confidence of 70\% or above mapped to expert ratings of 3 (neutral) or higher.
Over the Human-AI Spec-Solution Co-Optimization process, the main set of refinements related to the team's objective, under the broad category of \texttt{Problem Formalization}.
The key challenge was that patients almost always had multiple factors, both medical and operational, that interacted with each other in complex ways to contribute to prolonged LOS.
To improve consistency of both the extracted reasons and their assigned scores, successive iterations refined the problem definition, from focusing on factors that prolonged LOS by a minimum number of hours to providing more detailed instructions on how to account for interactions between factors.
While these made the AI agent's reasoning more systematic and mathematical, these definitions did not align with the implicit rating system used by clinical experts.
Only through repeated discussions did the team realize that the most clinically-aligned objective was patient flow, i.e., the efficiency of a patient's progression through individual hospital workstreams, not overall LOS, which requires understanding interactions between different factors and workstreams.
These changes to the objective naturally required corresponding updates to prompt tuning under \texttt{Model Learning}.

\begin{figure}[ht!]
\centering
\includegraphics[width=\linewidth]{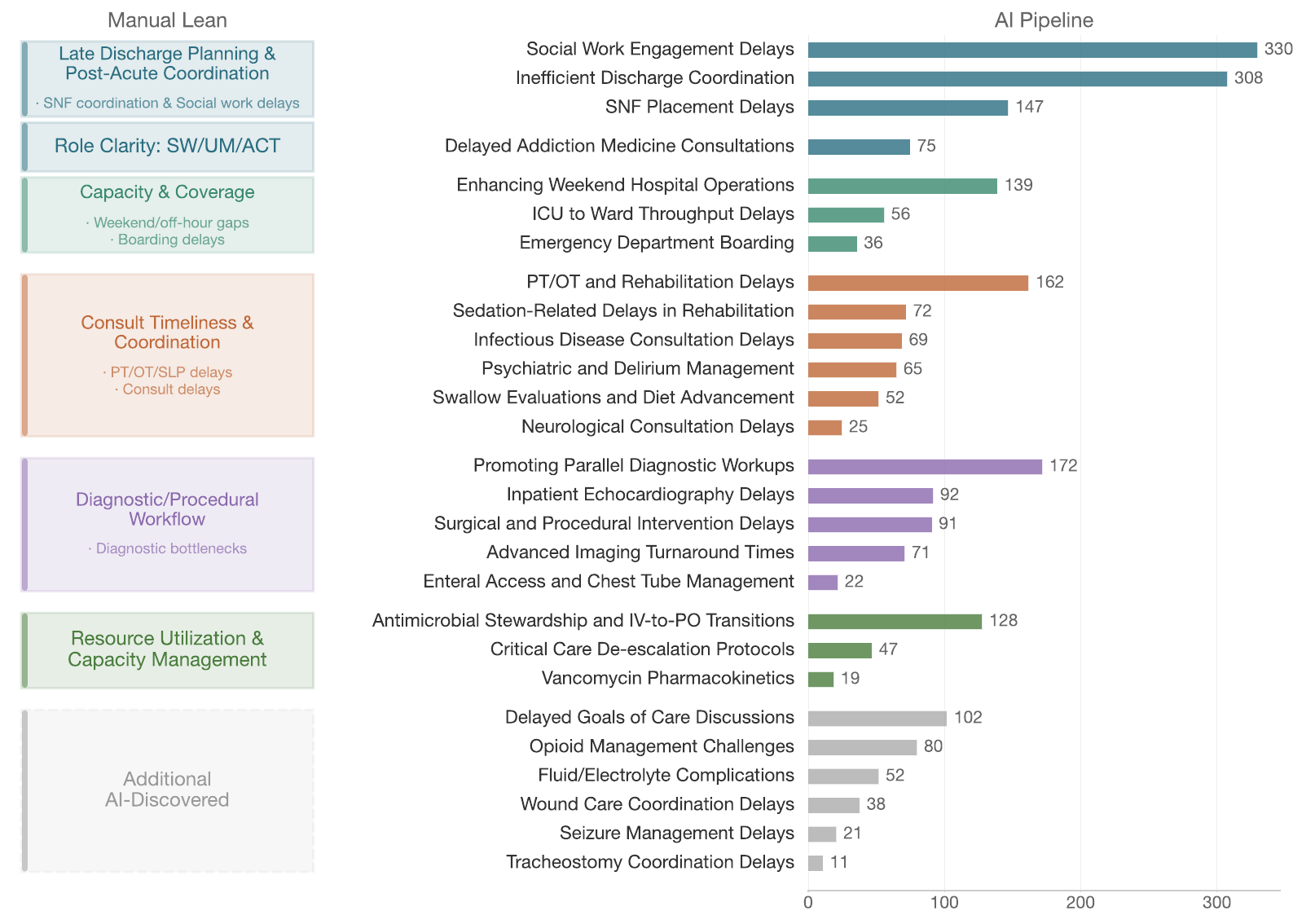}
\caption{Comparison of manual Lean analysis and AI pipeline results for prolonged LOS.
Left: the six categories identified by prior Lean analysis that primarily relied on manual expert review (25 patients, $\sim$100 person-hours), with sub-findings listed below each category.
Right: AI-identified modifiable care-flow gaps among 500 scaled-up encounters, of which 478 had at least one tagged factor. Themes are color-coded by their corresponding Lean category. The AI pipeline surfaced factors in all six Lean categories.
The ``Additional AI-Discovered'' group contains themes identified exclusively by the AI pipeline, absent from the manual analysis.
Bar length indicates unique patient encounters; a single encounter may contribute to multiple themes.
Factors that matched no theme do not contribute to any bar.
The complete list of all 27 themes is provided in Supplementary Table~S1.}
\label{fig:los_distribution}
\end{figure}

Separately, the team refined \texttt{Model Validation}.
Full chart review was impractical, so reviewers instead assessed the AI's extracted reasoning and evidence.
To reduce bias in this targeted review, the pipeline was iteratively modified to surface increasing amounts of information: first explanations and quotes, then process improvement suggestions, and finally both supportive and contrary evidence for each factor.
The review interface was revised in parallel to surface these outputs and help reviewers navigate charts more efficiently.
Additional refinements included splitting factor extraction from scoring, upgrading from GPT-5 Mini to Claude Opus 4.5, and progressively recruiting higher-expertise annotators; when several specifications changed at once (v5), validation returned briefly to the low-cost single-reader screen before review effort increased again (Figure~\ref{fig:los_heatmap}).
After the held-out evaluation, we scaled up the pipeline to 500 encounters across the five target DRGs, of which 478 were successfully labeled with matching tags.
After clustering the extracted contributors into 27 themes, we compared them against the six categories from the hospital's prior manual Lean analysis (Figure~\ref{fig:los_distribution}).
All six Lean categories were recovered by the AI pipeline, including discharge planning, consult timeliness, capacity constraints, diagnostic workflow bottlenecks, resource utilization, and role clarity gaps.
The AI pipeline additionally identified six themes absent from the manual analysis, such as delayed goals of care discussions, opioid management challenges, and fluid/electrolyte complications.
Beyond recovering and extending the Lean findings, the pipeline reviewed substantially more patient charts, provided a quantitative ranking of contributors, and generated auditable reasoning traces linked to individual encounters.

\subsection{Case Study 2: Unplanned 30-Day Readmissions}

\begin{landscape}
\begin{figure}[ht!]
\centering
\includegraphics[width=\linewidth]{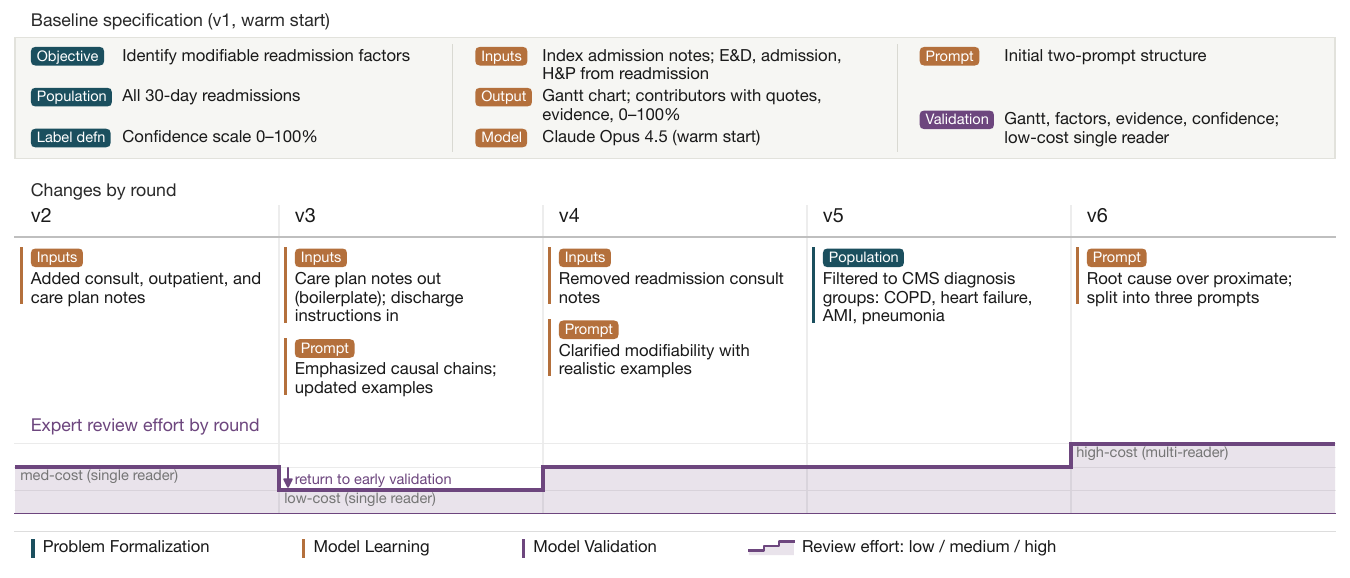}
\caption{Pipeline refinements across readmission optimization rounds.
The top panel shows the initial (v1) specification, which was warm-started from the LOS analysis.
Each column below lists the components changed in one round, colored by AI/ML development stage (Table~\ref{tab:generalized_pipeline}); components not listed were carried forward unchanged.
The bottom panel tracks expert review effort by round.
Full per-round specifications and the feedback behind each change appear in the Supplementary Information.
}
\label{fig:readm_heatmap}
\end{figure}
\end{landscape}

The ZSFG QI team applied the Human-AI Spec-Solution Co-Optimization to develop an AI pipeline for identifying top modifiable contributing factors to 30-day unplanned readmissions.
The progression of the problem specifications is summarized in Figure~\ref{fig:readm_heatmap}.
We note that due to hospital resourcing, this case study was conducted after the LOS case study, so the group was able to leverage existing prompts and high-level problem specifications as a ``warm start'' \citep{Ash2020-ik}.
This case study is likely more representative of the types of changes that other hospitals may make if they were to conduct a similar procedure, as they can build on lessons and artifacts from this work.

The final AI pipeline was evaluated on 210 candidate factors extracted across 44 held-out encounters, for which the domain experts provided a total of 290 annotations.
On the 20 factors rated by all five reviewers, the within-one-point agreement rates were 65.0\% for LLM-rater versus 72.5\% for inter-rater agreement, and the difference was not statistically significant (95\% CI for the difference $-19.0$ to $3.0$).
For the full set of metrics compared, we refer readers to Supplementary Table~S8.
Across the full set of factor annotations, agreement within one Likert point was 71.0\% (95\% CI 65.2--76.7\%) (Figure~\ref{fig:pipeline_eval} right;
Supplementary Table~S7). %
The calibration curve confirmed that higher LLM confidence corresponded to higher expert ratings, with the 50--59\% confidence bin corresponding to the lowest bin whose estimated mean expert rating was 3 (neutral) or higher.

The two major themes that emerged in the Human-AI Spec-Solution Co-Optimization process were in \texttt{Problem Formalization} and \texttt{Model Learning}.
Updates to the \texttt{Model Validation} specifications were much more limited, since the validation process could mostly mimic that used in the LOS case study.

Under \texttt{Problem Formalization}, the major change was to narrow the ``Population.''
Initially, the QI team was planning to target all 30-day unplanned readmissions, but many of these cases were difficult to validate or less clinically relevant.
Examples included COVID-19 cases and patients with rarer diseases where the healthcare providers were not experts.
Consequently, the team decided to focus on diagnosis groups highlighted by the Centers for Medicare \& Medicaid Services (chronic obstructive pulmonary disease [COPD], heart failure [HF], acute myocardial infarction [AMI], and pneumonia), as the team had sufficient expertise to validate such cases.

\begin{figure}[ht!]
\centering
\includegraphics[width=\linewidth]{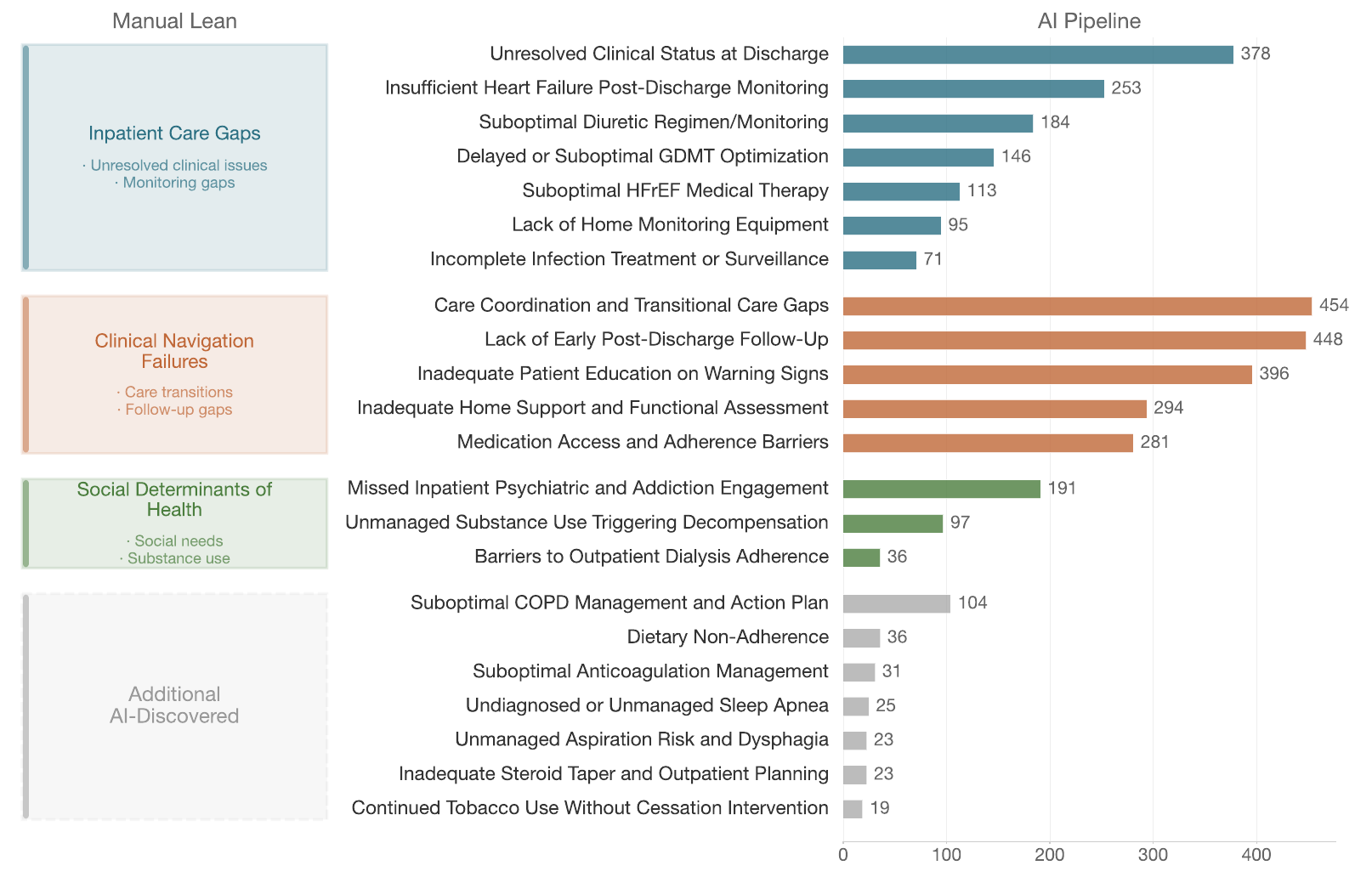}
\caption{Comparison of manual Lean A3 analysis and AI pipeline results for 30-day unplanned readmission.
Left: the three categories identified by prior Lean A3 analysis that primarily relied on manual expert review (25 patients, $\sim$100 person-hours), with sub-findings listed below each category.
Right: AI-identified modifiable care-flow gaps from the 500 scaled-up encounters, of which 484 had tagged factors. Themes are color-coded by their corresponding Lean category. The AI pipeline surfaced factors in all three Lean categories.
The ``Additional AI-Discovered'' group contains themes identified exclusively by the AI pipeline, absent from the manual analysis.
Bar length indicates unique patient encounters; a single encounter may contribute to multiple themes.
Factors that matched no theme do not contribute to any bar.
The complete list of all 22 themes is provided in Supplementary Table~S2.}
\label{fig:readmission_distribution}
\end{figure}

Under \texttt{Model Learning}, the team also had to iteratively modify ``Estimator Inputs.''
Unlike LOS, where all relevant information lived in a single hospitalization, readmission analysis required tracing causal chains across the index admission, outpatient visits, and the readmission itself.
The team initially provided the AI agent with only the admission note from the index encounter and the admission note and H\&P from the readmission, but the LLM and annotators lacked sufficient clinical context to make confident assessments.
Successive iterations added consult notes, outpatient notes, care plan notes, and discharge instructions.
Not all note types proved useful: care plan notes were mostly templated boilerplate that added noise rather than signal, and were removed.
Finally, due to the complexity of multi-encounter charts, the team had to further split up prompts to avoid API timeouts.

We then scaled up the final pipeline to 500 encounters across the target CMS diagnosis groups (COPD, HF, AMI, and pneumonia), of which 484 were labeled with matching tags and which, after clustering, yielded 22 themes (Figure~\ref{fig:readmission_distribution} right).
The hospital's prior manual Lean A3 analysis, which focused on readmissions in HF patients alone due to resource constraints, had identified three broad categories: inpatient care gaps, clinical navigation failures, and social determinants of health (Figure~\ref{fig:readmission_distribution} left).
In contrast, the AI pipeline not only recovered these three categories but also identified seven additional themes absent from the manual analysis.
Results are similar when the AI-extracted themes are restricted to HF patients only, except that the themes relating to HF increase in prevalence (Supplementary Table~S3).
Some of the additional themes (e.g., suboptimal COPD management, unmanaged aspiration risk) largely disappear in the HF subgroup and thus reflect the broader case mix rather than discovery within the Lean population, while others relevant to HF (e.g., dietary non-adherence, suboptimal anticoagulation management) persist.
These quantitative rankings can help QI teams prioritize which factors to address first.

\section{Discussion}

Quality improvement programs play a critical role in improving hospital processes and outcomes, but most face severe resource constraints that limit the scope and frequency of their analyses.
While there is extensive work on implementing AI as part of specific hospital QI interventions \citep{Hong2020-vb, Martinez2022-aw, Adams2022-rm}, this work addresses the hypothesis-generation phase of QI---the critical first step of identifying candidate factors for potential intervention---demonstrating that AI can systematically surface hypotheses aligned with expert intent from clinical data at scale in a manner that is adaptable to other settings, more readily reproducible within the study setting, and auditable through chart-linked evidence artifacts.
We show that the ``fuzzy'' exploratory, expert-driven process at the heart of QI hypothesis generation can be systematized into a formal AI pipeline without sacrificing the clinical judgment that makes it effective.
Furthermore, AI can complement qualitative QI methods by leveraging a greater volume of clinical data, producing candidate hypotheses linked with evidence from patient charts.
These advancements can also help hospitals progress to become Learning Health Systems \citep{Etheredge2007-on, Institute_of_Medicine2013-uf}, as AI pipelines can continuously monitor for new potential contributors to target.
Our experiments show that our AI-for-QI pipelines, iteratively optimized and retrospectively applied at a large urban safety-net hospital, achieved a $\ge70\%$ agreement rate with domain experts for two QI metrics: length of stay (LOS) and 30-day unplanned readmissions.
In terms of cost, Human-AI Spec-Solution Co-Optimization is substantially more scalable.
An institution adopting this framework only needs a one-time investment in setting up the computational infrastructure and expert time to iterate and validate the LLM pipeline components.
To this end, we have open-sourced the code for Human-AI Spec-Solution Co-Optimization and provided step-by-step instructions and tutorials; the primary requirements to run the pipeline are a dataset with patient charts and access to a HIPAA-compliant LLM.
Furthermore, we found that expert time needed was minimal for the two presented case studies: less than 5 hours were spent per case study reviewing $\le$20 charts during the co-optimization phase with assistance from AI, and around 10--12 person-hours per case study were needed to review 52 (LOS) and 44 (readmission) encounters during the evaluation phase. %
Once tuned, the pipeline analyzed the full scale-up cohort in roughly 30 minutes.
These figures cover expert review and inference only and exclude the one-time effort of developing the pipeline, which adopting institutions need not repeat.
In contrast, each prior Lean analysis required approximately 100 person-hours for 25 patients and would require a similar amount of time if conducted again.

The key methodological contribution of this work is the development of a formal framework for translating ``fuzzy'' QI analyses into a concrete AI pipeline, a challenge that has until now limited the successful use of LLMs in quality improvement science.
In particular, unlike typical settings considered in the AI literature, QI hypothesis generation is uniquely challenging because the very goal of a hospital system's strategic planning process is to translate high-level hospital priorities and KPIs into concrete QI analyses and initiatives.
Despite the growing number of semi-structured tools in QI hypothesis generation, this translation process is inherently messy, because teams must learn as they go and many critical decisions require expert judgment, often made implicitly.
Transforming QI analyses into an AI pipeline requires formalizing the process while maintaining its critical exploratory components.
The key insight in this work is that we can do this if we map the process of aligning an AI pipeline with the QI team's goals to steps of the AI/ML development process where we must not only tune the AI pipeline itself but also each step's specifications.
Through this framing, we optimize the AI pipeline and the problem's high-level specifications simultaneously with input from both the QI team and the AI agent.
This process, Human-AI Spec-Solution Co-Optimization, is key to ensuring that the trained AI pipeline aligns with a QI team's priorities and resource constraints.
In an ablation analysis comparing a generic prompt to one refined through this co-optimization process, factors extracted with the generic prompt covered 26 of the 27 final LOS themes when tagged against the final taxonomy (Supplementary Information), which suggests that the main advantage of co-optimization lies in the specificity, ranking, alignment with local priorities, and evidentiary support of the extracted factors rather than in recovering broad themes.
Furthermore, this general framework is likely more broadly applicable, beyond the specific task of QI hypothesis generation.

Given the centrality of AI-driven insights in the QI methodology we propose, it is important to consider how institutional governance can promote responsible AI use and monitoring in the QI context~\citep{Aldosari2025}.
Our findings suggest that large language models can recover and extend factors identified by human QI teams; nevertheless, human oversight is always required in our framework to refine the specification for chart review and interpret the pipeline outputs.
The framework's intended use is hypothesis generation by hospital QI teams from retrospective records: its inputs are clinical notes from completed encounters, its end users are QI analysts and clinicians, every output is reviewed by experts before informing prioritization, and no output reaches patient care or real-time clinical decision-making.
Institutional training on the strengths and weaknesses of AI-based chart review and fostering cross-disciplinary QI teams can ensure that this oversight is effective.
When translating the discovered hypotheses to hospital QI efforts, our pipeline improves auditability by yielding full documentation of how LLMs were prompted to produce the results; however, this benefit hinges upon QI teams reporting this iterative process for stakeholders to review.
The pipeline also naturally prioritizes factors that LLMs are better at identifying and extracting from clinical notes.
For this reason, we recommend that teams use this framework as one tool among many in the QI toolbox.

The primary limitation of this framework is that it is a workflow-support tool for hypothesis generation, not hypothesis validation; the factors identified by the pipeline are candidates for intervention and the causal effects of such interventions depend heavily on their implementation.
As such, this framework does not provide any causal analysis, in line with other tools for QI hypothesis generation \citep{Bennett2025-tp}.
Causal estimation remains an open problem in this setting, since QI teams often propose novel interventions with little or no precedent, leaving insufficient data to estimate their effects (e.g., due to positivity violation); we refer readers to emerging research in this area that formalizes causal inference in the QI setting \citep{vossler2026expertguidedgcomputationlargelanguage}.
As such, the value of the framework is that it systematically surfaces hypotheses that would otherwise require extensive manual chart review and can provide quantitative evidence for which factors to prioritize based on their prevalence in patient charts.

A related limitation is that through this Human-AI Spec-Solution Co-Optimization process, teams settled on AI-supported validation workflows, as this provided a favorable balance between practical feasibility/scalability and the potential risks of anchoring and automation bias.
Reviewers rated factors that the AI had already surfaced and viewed the AI's extracted reasoning together with supporting and contrary evidence, rather than performing independent chart review.
As such, the final validation metrics measure alignment between the pipeline and expert judgment, but not completeness of hypothesis generation (e.g., recall and false negative rates).
Human experts were encouraged to share additional free-text feedback on factors missed by the pipeline, but no such feedback was submitted in practice given resource constraints.
In future work, it would be valuable to conduct prospective evaluations comparing factors generated by human experts alone versus those generated through this iterative human-AI pipeline.
It would also be important to assess fairness of the AI pipeline to ensure equitable performance across groups \citep{Feng2023-ua, Zack2024-ft}.

Both case studies presented in this work were conducted retrospectively at a single urban safety-net hospital (ZSFG) and specifications/prompts were co-optimized with a particular model.
As such, the framework's performance may differ when run in a different hospital with different LLMs, particularly depending on the completeness, size, and diversity of patient charts.
Nonetheless, the pipeline code and all prompt versions are open-source, available for QI teams at other hospitals to adopt and improve upon.
We expect many of the LLM instructions to generalize to other institutions with minor modifications, offering a useful ``warm start'' for other hospitals conducting the Human-AI Spec-Solution Co-Optimization process.
Through the co-optimization process, expert teams will also be able to evaluate if the available clinical documentation is sufficient and, if not, decide if more data needs to be gathered or if  the QI target needs to be refined.

This work illustrates how generative AI can support the early hypothesis-generation work of hospital QI.
Human-AI Spec-Solution Co-Optimization formalizes an otherwise iterative, qualitative, and fuzzy process, preserves expert judgment, and creates reusable artifacts that can be inspected, rerun, and adapted.
While our case studies focused on prolonged LOS and 30-day unplanned readmissions, the same framework may apply to other QI settings where teams must clarify an underspecified analytic task before designing an intervention.
Our institution is now planning to implement prospective QI initiatives based on the factors identified using this AI-for-QI framework.
Following the Lean Healthcare framework, outputs from this pipeline correspond to the problem-analysis stage of an improvement cycle (e.g., the left side of an A3 report or the Plan step of a Plan-Do-Study-Act (PDSA) cycle) \citep{shook2008managing, langley2009improvement}.
The next stage is to design, refine, and evaluate QI interventions through iterative PDSA cycles and prospective studies such as interrupted time series, stepped-wedge rollouts, and statistical process control charts \citep{taylor2014systematic, benneyan2003statistical, portela2015how}, with ineffective interventions de-implemented.
Through the improvement cycles, these initiatives will ultimately determine whether the QI factors identified by the framework can be translated into actionable interventions to improve hospital operations and quality of care.

\section{Methods}
\label{sec:methods}

\subsection{QI Hypothesis Generation Framework}

We formalize QI hypothesis generation by mapping it onto the steps and sub-steps of the classical AI/ML development process---\texttt{Problem Formalization}, \texttt{Model Learning}, and \texttt{Model Validation} (Table~\ref{tab:generalized_pipeline})---so that the LLM prompts and the high-level specifications of the problem and solution become natural-language-valued hyperparameters to be tuned.
Initially, these specifications are vague.
Indeed, both of the presented case studies in this work began with specifications of the following form:
\begin{enumerate}
\item \texttt{Problem Formalization}: ``Identify operational factors in patient charts that led to $X$'' where $X$ is ``prolonged LOS'' or ``increased risk of unplanned 30-day readmission.''
\item \texttt{Model Learning}: ``An AI pipeline should analyze discharge summaries to extract three possible reasons that led to $X$ and rate the reasons on a Likert scale from 1--3.''
\item \texttt{Model Validation}: ``Experts will score the extracted reasons on a Likert scale from 1--3 and concordance is defined as how often the AI- and expert-assigned scores match.''
\end{enumerate}
These specifications must be tuned until the process of QI hypothesis generation and the validation results are fully documented and re-executable within the study context; this means the specifications must be precise enough that teams can reproduce data extraction, execution of the AI pipeline, and the annotation and validation process.
They must also be limited to only specifications that the team considers ``valid,'' or meaningful to external stakeholders.
Tuning problem definitions and validation metrics is statistically permissible under this framing, because the final pipeline is evaluated on held-out encounters after the specifications have been fixed.
Once candidate factors have been identified, they can be audited and interpreted alongside the specifications and validation process.

Beyond providing a formal framework, this mapping has a practical benefit: it equips teams with the precise terminology needed to characterize and address technical challenges that would otherwise be difficult to articulate.
For instance, a vague concern like ``we don't have the resources to fully review AI outputs'' can be formalized as a challenge within the \texttt{Model Validation} step, where one only has imperfect reference standards (or silver-standard labels).
This then provides pointers to not only
statistical corrections such as prediction-powered inference (PPI) and Bayesian inference \citep{Angelopoulos2023-tv, Vossler2026-ii} but also simpler natural-language solutions, such as designing annotation interfaces that surface opposing evidence to reduce reviewer bias.
This mapping also makes clear what distinguishes QI hypothesis generation from prior work: while only one or two rows of Table~\ref{tab:generalized_pipeline} lacked clarity in prior settings \citep{Kothari2026-zo, Feng2026-ip}, nearly \textit{every} row is underspecified in QI hypothesis generation.

\subsection{Human-AI Spec-Solution Co-Optimization}

Once the QI hypothesis-generation task is formalized, learning an AI-for-QI pipeline becomes an optimization problem over the space of natural-language descriptions that parameterize each step.
The goal is to refine these specifications so that the steps are mutually consistent and aligned with clinical objectives, resource constraints, and AI capabilities.
Because changing one specification often affects others (for example, redefining the objective may require different validation criteria), this optimization must be performed jointly, with input from both the domain expert and AI agent.

In practice, the exact implementation of this co-optimization procedure varies based on the expertise makeup of the team and its resourcing constraints.
In this work, the two key constraints were that (i) expert time was highly limited and (ii) a full review of a patient chart was prohibitively time-consuming.
To address these constraints, we implemented two solutions.
First, we designed a PHI-compliant web interface to assist domain experts in reviewing AI outputs and reasoning (Figure~\ref{fig:workflow_ui} bottom).
This interface was composed of two main panels: a Gantt chart abstracting the patient journey and a list of potential contributing factors with AI-extracted pointers to key pieces of supporting and contrary evidence.
We updated this interface throughout the refinement process, for instance by increasing the amount of extracted evidence and LLM reasoning surfaced to mitigate reviewer bias.
No formal usability instrument or human-factors framework was applied to the review interface; its design evolved across the co-optimization rounds in response to reviewer feedback.
Second, rather than collecting all annotations upfront, we used an \textit{online} (incremental) process, iteratively annotating more data with silver-standard labels and progressively improving label quality to support model validation (Figure~\ref{fig:workflow_ui} top).
In both case studies, the process generally began with a single data scientist iterating with the AI agent, expanded to include a single clinician, and concluded with a full group alignment meeting before the final validation round.
Model choice followed the same escalation: the data scientist initially selected a smaller model to limit cost, clinician review found its clinical reasoning insufficient, and after informal testing of larger models the team fixed Claude Opus 4.5 for the final rounds of both case studies.
Model changes were driven by expert judgment rather than head-to-head benchmarking, and no change occurred after the evaluation rounds.

\subsection{Statistical Inference: Calibration and evaluation}

As all iterative refinements to the pipeline constitute a form of training on the data used during optimization, the evaluation and calibration of the final AI pipeline must be conducted on encounters held out from those refinement rounds to ensure valid statistical inference.

For the LOS case study we assembled the full eligible cohort of 2,029 encounters across five target diagnosis groups, and for the readmissions case study we assembled 777 encounters from the hospital readmission registry across four CMS diagnosis cohorts.
Encounters were randomly selected for every refinement and evaluation round, and the evaluation encounters were held out from the iterative refinement rounds; note that encounters reviewed in prior Lean analyses were not part of the tuning or evaluation cohorts.
After validation, each pipeline was scaled up to 500 sampled encounters.
Of these, 478 (LOS) and 484 (readmission) had at least one factor at or above the confidence threshold for clustering, and these encounters form the denominator of the theme analysis and are characterized in Supplementary Tables~S4--S6.
Early-stage QI hypothesis generation does not have a natural gold-standard label: the output is a set of candidate factors for prioritization, not a known diagnosis or a prospectively verified causal effect.
Accordingly, we evaluated the learned AI pipeline as a workflow-support method rather than as a clinical intervention.
The evaluation separates two questions.
First, on expert-reviewed held-out encounters, does the pipeline assign ratings to candidate factors that are concordant with expert judgment?
Second, when the fixed pipeline is applied to the larger held-out cohort, do the resulting factor categories recover the hospital's prior manual Lean findings?

To evaluate the first question, the pipeline assigned a confidence score to each candidate factor, which was then compared with the rating assigned by a human reviewer following the tuned \texttt{Model Validation} specifications.
We report exact agreement and within-one-point agreement on the prespecified five-point Likert scale, with the latter reflecting meaningful alignment in a subjective hypothesis-generation task.

Agreement was quantified in two complementary analyses.
The first set of agreement metrics was evaluated using every expert rating provided in a held-out evaluation set, where each factor's confidence score was mapped to the five-point scale with prespecified confidence bins.
For this analysis, we report exact and within-one-point agreement, mean absolute error, weighted kappa, as well as Spearman and Kendall rank correlations for the raw un-binned scores;
95\% confidence intervals were calculated with a bootstrap that resamples candidate factors, matching the resampling unit of the multi-reviewer comparison below; encounter-clustered intervals are reported in the Supplementary Information as a sensitivity analysis for the clustering of factors within patients.
The second set of analyses compares agreement rates between the AI pipeline and expert raters versus agreement rates between expert raters; this comparison was evaluated using only items rated by multiple reviewers.
On this subset we report pooled pairwise inter-rater agreement, LLM-rater agreement with the pipeline treated as an additional rater, and their difference, with bootstrap confidence intervals.
We note that \texttt{Model Validation} specifications are allowed to provide human reviewers AI assistance during the annotation process, \textit{as long as the validation process is fully specified.}
This ensures full transparency in how validation was conducted and improves the method's reproducibility.
Furthermore, to mitigate the risk of automation bias, the specifications were progressively modified to make AI assistance more objective, such as by surfacing both supportive and contrary evidence in the annotation interface.
Thus, the resulting evaluation metric represents a careful and deliberate tradeoff between feasibility and unbiasedness, and results should be interpreted accordingly.

The second question measures consistency with prior QI work at the hospital.
After the specifications were fixed, the final pipeline was applied to the larger held-out cohort for each metric.
We then compared the resulting factor categories with categories previously identified through manual Lean analyses at ZSFG as well as additional high-frequency factors.
This comparison evaluates whether the AI-assisted workflow recovers and extends prior QI work at the hospital; it does not establish that the identified factors are causal, modifiable, or generalizable to other hospitals.

We also report the number of encounters processed, runtime, and evidence artifacts produced by the final pipeline as descriptive properties of the workflow.
These quantities characterize the scale and inspectability of the resulting analysis, but they are not treated as independent validation endpoints.
Causal validity, intervention feasibility, and clinical impact require downstream prioritization, intervention design, and prospective evaluation.

The AI pipeline outputs a confidence score (0--100) for each extracted QI factor, but these raw scores are not necessarily calibrated to expert judgment.
As an exploratory post-hoc evaluation of calibration, we binned the LLM confidence scores into intervals and computed the mean expert rating within each bin, along with 95\% confidence intervals.
Because the 0--100 confidence scale does not map directly onto the 1--5 Likert scale used by human annotators, binning allows us to estimate, at each confidence level, the average rating a human reviewer would assign.
We used 10-percentage-point bins at higher confidence values and combined lower confidences into a single bin to keep within-bin sample sizes large enough for reliable mean estimation.
Bin edges were fixed in each case study's pipeline before the held-out evaluation and differ slightly between the two case studies because the distributions of LLM confidences differed.
Thus, calibration curves reported in Section~\ref{sec:results} above summarize how the mean expert rating varies with the confidence score.
We emphasize that our evaluations of expected calibration error should be viewed as an exploratory, post hoc comparison of the confidence score with expert ratings.

\subsection{Clustering and Tagging Extracted Factors}

We conducted a three-step procedure to identify themes across the factors extracted by the AI pipeline during the final scaled-up runs.
First, we used BERTopic to do unsupervised clustering \citep{Grootendorst2022-it}, in which each factor was embedded using \texttt{all-MiniLM-L6-v2} and clustered using HDBSCAN.
The learned clusters were then jointly interpreted by an LLM and a human expert (J.F., who was not involved in the prior Lean analyses) to generate cohesive, clinically relevant themes.
Finally, each AI-extracted factor was individually tagged with zero or more relevant themes by an LLM, and an independent reviewer checked the themes assigned to 30 randomly sampled factors per case study (Supplementary Information, ``Theme Tag Validation'').
Clustering settings and the handling of outliers, merged topics, and factors with several or no themes are described in the Supplementary Information section ``Implementation Details''.
These final tags were collated to compute the prevalence of the different themes.

\subsection{Reporting standards}
This study was reported following SQUIRE 2.0, treating the AI-assisted hypothesis-generation workflow as the initiative under study \citep{ogrinc2016squire}, and TRIPOD-LLM, under which it is an evaluation of a large language model in a dedicated healthcare setting performing information-extraction and classification tasks \citep{gallifant2025tripodllm}.
We additionally adapted items from the DECIDE-AI guideline relevant to human-AI evaluation (intended use, system description and output presentation, human factors of the review interface, and human--computer agreement), as the original guideline targets early-stage live clinical evaluation \citep{vasey2022decideai}.
TRIPOD-LLM items specific to de novo model development or fine-tuning are not applicable because the pipeline prompts general-purpose models without any parameter updates, and the completed checklist marks them as such.
Completed SQUIRE 2.0 and TRIPOD-LLM checklists appear in the Supplementary Information (``Reporting Checklists'').

\section{Data availability}
The patient data that support the findings of this study are not publicly available because they consist of electronic health record data, which constitute protected health information. To support reproduction of this analysis at other institutions, the Supplementary Information provides the full prompt text, the per-iteration specification sheets, and the learning-process details for both case studies; de-identified or synthetic examples are also included in the associated open-source software package.

\section{Code availability}
The complete AI-for-QI pipeline is publicly available at \url{https://github.com/jjfenglab/llm-qi} under a GPL-3.0 license.
The repository contains the full pipeline code for both case studies, including all prompt versions, the PHI-compliant web interface for expert review and annotation, the statistical analysis code, and tutorials for running the Human-AI Spec-Solution Co-Optimization procedure on a hospital's own data.

\section{Acknowledgements}
The PROSPECT lab (P.V., J.F., L.Z., J.M.) thanks the Zuckerberg Priscilla Chan quality improvement fund, via the San Francisco General Foundation, for funding this project.
R.G. reports support from Award P30HS029738 from the Agency for Healthcare Research and Quality (AHRQ) and the Patient-Centered Outcomes Research Institute (PCORI). The content is solely the responsibility of the authors and does not necessarily represent the official views of AHRQ or PCORI.
H.K. reports grant funding to the University of California, San Francisco, from the California Department of State Hospitals and the San Francisco Department of Public Health.
The authors thank the UCSF AI Tiger Team, Academic Research Services, Research Information Technology, and the Chancellor's Task Force for Generative AI for technical support related to the use of the Versa API gateway.
The funders had no role in study design, data collection, analysis, interpretation, or the decision to submit for publication.

\section{Author contributions}
P.V., J.F., L.Z., H.K., C.R., and C.P. conceived the study. P.V. and J.F. developed the methodology and software, with V.S. contributing to the web interface. J.F. led the annotation efforts and data analysis, with contributions from P.V., V.S., R.G., H.K., C.R., J.M., C.P., and L.Z. R.G., H.K., D.F., C.R., A.O., J.M., S.E., and C.P. provided clinical and quality-improvement expertise, contributed to annotation and validation, and reviewed and edited the manuscript. P.V., J.F., and V.S. wrote the manuscript. L.Z. and J.F. supervised the work. All authors read and approved the final manuscript.

\section{Competing interests}
H.K. serves as an advisor with equity for Amae Health, outside the submitted work, and is an unpaid Board Member for the San Francisco General Hospital Foundation. S.E. serves in unpaid volunteer roles on the boards of the Institute for Healthcare Improvement (IHI), the California Hospital Association, Health Alliance, and the San Francisco General Hospital Foundation. The remaining authors declare no competing financial or non-financial interests.

\section{Ethics statement}
This study was conducted as a quality improvement initiative at Zuckerberg San Francisco General Hospital and Trauma Center. It was approved by the University of California, San Francisco Institutional Review Board (protocol 22-36613) as well as Zuckerberg San Francisco General Hospital. Individual informed consent was waived given the retrospective nature of the analysis. All data were stored and analyzed within institutional systems in compliance with HIPAA regulations.

This work utilized UCSF's secure Generative AI platform, Versa, which interfaces with HIPAA-eligible LLM services through Azure and Amazon Web Services.
Under institutional enterprise configurations, vendor endpoints do not retain user prompts or outputs for continued model training or monetization.
Access to the system was strictly regulated by UCSF's secure infrastructure, which enforces rigorous access controls, network isolation, and logging under a shared-responsibility model.
The deployment of these tools operates under the continuous oversight of UCSF IT, alongside internal Privacy, Legal, and AI Governance processes, ensuring that appropriate safety, business associate frameworks, and protected health information (PHI) compliance reviews are in place.

\bibliography{refs}

\clearpage
\begin{center}{\Large\bfseries Supplementary Information}\end{center}
\setcounter{section}{0}
\setcounter{table}{0}
\setcounter{figure}{0}
\renewcommand{\thesection}{S\arabic{section}}
\renewcommand{\thetable}{S\arabic{table}}
\renewcommand{\thefigure}{S\arabic{figure}}
\renewcommand{\tablename}{Supplementary Table}
\renewcommand{\figurename}{Supplementary Fig.}
\raggedbottom

\section{Implementation Details}

The AI pipeline was implemented in Python.
The final pipelines used Claude Opus 4.5 (Anthropic; \texttt{us.anthropic.claude-opus-4-5-20251101-v1:0}) for patient-journey extraction, factor extraction, and confidence scoring; earlier iterations used GPT-5 Mini and GPT-5 (OpenAI).
Grouping the extracted factors into themes used Claude Haiku 4.5 (Anthropic; \texttt{us.anthropic.claude-haiku-4-5-20251001-v1:0}).
OpenAI models were accessed through UCSF's Versa API gateway and Anthropic models through AWS Bedrock, both covered by institutional agreements for handling protected health information.
Inference used temperature 1 with outputs returned as structured JSON and validated against a fixed schema. %
Stage-specific maximum output lengths were used (5{,}000--10{,}000 tokens for extraction and 1{,}000 for theme tagging); each encounter was processed once per round, and the final evaluated and scaled-up runs were performed in March--April 2026.
Because outputs were sampled at temperature 1 and each encounter was processed once, extracted factors, confidence scores, and theme prevalences may vary between runs, and run-to-run stability was not evaluated.
A PHI-compliant web interface was developed for expert review and annotation of AI-extracted factors (see Methods in the main text).
The complete pipeline code is publicly released as open-source software.

Only factors with a confidence score of at least 70 (LOS) or 50 (readmission) entered the clustering step.
Within BERTopic, factors were embedded with \texttt{all-MiniLM-L6-v2}, reduced with UMAP (10 neighbors, 5 components, minimum distance 0, cosine metric), and clustered with HDBSCAN (minimum cluster size 10, minimum samples 3), and BERTopic capped the number of topics at 30 by merging the most similar clusters when HDBSCAN found more.
Because some topics were much more specific than others, the human expert merged overly specific topics that shared a clinical domain or would be addressed by the same QI initiative, yielding 27 LOS and 22 readmission themes.
Every factor at or above the confidence threshold, including factors that HDBSCAN left unclustered, was then tagged with zero or more themes, so no outliers were removed.
Of the 1,643 LOS factors, 1,111 (68\%) received more than one theme, 496 (30\%) received one, and 36 (2\%) received none.
Of the 2,018 readmission factors, 1,931 (96\%) received more than one theme, 79 (4\%) received one, and 8 (0.4\%) received none.
Nine of the factors without a theme (1 LOS and 8 readmission) reflect tagging calls that did not return a usable response, and the remaining 35 LOS factors were narrowly specific issues that matched no theme.
Each factor counts toward every theme it received, and a theme's prevalence counts the encounters with at least one factor tagged with that theme.

\section{Full Theme Distribution for LOS Case Study}

Table~\ref{tab:los_all_themes} presents the complete set of 27 AI-identified themes for prolonged LOS, including both unique encounter counts and total reason counts.
Themes are grouped by their correspondence to the six manual Lean analysis categories where applicable.
A single encounter may contribute to multiple themes.

\begin{table}[h]
\centering
\small
\caption{All 27 AI-identified themes for prolonged LOS with unique encounter counts and total reason counts, grouped by manual Lean category alignment. Themes within each category are sorted by encounter count.}
\label{tab:los_all_themes}
\begin{tabular}{llrr}
\toprule
\textbf{Lean Category} & \textbf{AI-Identified Theme} & \textbf{Encounters} & \textbf{Reasons} \\
\midrule
\multirow{3}{*}{\shortstack[l]{Discharge Planning \&\\Post-Acute Coordination}}
 & Social Work Engagement & 330 & 490 \\
 & Inefficient Discharge Coordination & 308 & 486 \\
 & Skilled Nursing Facility (SNF) Placement & 147 & 237 \\
\addlinespace
Role Clarity (SW, UM, ACT)
 & Delayed Addiction Medicine Consultations & 75 & 105 \\
\addlinespace
\multirow{3}{*}{Capacity \& Coverage}
 & Enhancing Weekend Hospital Operations & 139 & 151 \\
 & ICU to Ward Throughput & 56 & 63 \\
 & Emergency Department Boarding & 36 & 37 \\
\addlinespace
\multirow{6}{*}{\shortstack[l]{Consult Timeliness\\\& Coordination}}
 & PT/OT and Rehabilitation & 162 & 224 \\
 & Sedation-Related Delays in Rehabilitation & 72 & 83 \\
 & Infectious Disease Consultation & 69 & 83 \\
 & Psychiatric and Delirium Management & 65 & 94 \\
 & Swallow Evaluations and Diet Advancement & 52 & 63 \\
 & Neurological Consultation & 25 & 27 \\
\addlinespace
\multirow{5}{*}{\shortstack[l]{Diagnostic/Procedural\\Workflow \& Turnaround}}
 & Promoting Parallel Diagnostic Workups & 172 & 244 \\
 & Inpatient Echocardiography & 92 & 115 \\
 & Surgical and Procedural Interventions & 91 & 120 \\
 & Advanced Imaging Turnaround Times & 71 & 86 \\
 & Enteral Access and Chest Tube Management & 22 & 27 \\
\addlinespace
\multirow{3}{*}{\shortstack[l]{Resource Utilization \&\\Capacity Management}}
 & Antimicrobial Stewardship and IV-to-PO Transitions & 128 & 164 \\
 & Critical Care De-escalation Protocols & 47 & 55 \\
 & Vancomycin Pharmacokinetics & 19 & 20 \\
\midrule
\multirow{6}{*}{\textit{AI-only (not in Lean)}}
 & Goals of Care and Palliative Care Integration & 90 & 136 \\
 & Inpatient Opioid Use Disorder Management & 49 & 62 \\
 & Complications of Fluid Status and Electrolytes & 36 & 38 \\
 & Inpatient Pain Management & 29 & 37 \\
 & Inpatient Glycemic Control and Endocrine Support & 16 & 19 \\
 & Inter-Facility Repatriation & 12 & 15 \\
\bottomrule
\end{tabular}
\end{table}

\section{Full Theme Distribution for Readmission Case Study}

Table~\ref{tab:readmission_all_themes} presents the complete set of 22 AI-identified themes for 30-day unplanned readmission, including both unique encounter counts and total reason counts.
Themes are grouped by their correspondence to the three categories from the manual Lean A3 analysis where applicable.
A single encounter may contribute to multiple themes.

\begin{table}[h]
\centering
\small
\caption{All 22 AI-identified themes for 30-day unplanned readmission with unique encounter counts and total reason counts, grouped by manual Lean A3 category alignment. Themes within each category are sorted by encounter count.}
\label{tab:readmission_all_themes}
\begin{tabular}{llrr}
\toprule
\textbf{Lean Category} & \textbf{AI-Identified Theme} & \textbf{Encounters} & \textbf{Reasons} \\
\midrule
\multirow{7}{*}{Inpatient Care Gaps}
 & Unresolved Clinical Status at Discharge & 378 & 701 \\
 & Insufficient Heart Failure Post-Discharge Monitoring & 253 & 450 \\
 & Suboptimal Diuretic Regimen/Monitoring & 184 & 312 \\
 & Delayed or Suboptimal GDMT Optimization & 146 & 205 \\
 & Suboptimal HFrEF Medical Therapy \& Monitoring & 113 & 152 \\
 & Lack of Home Monitoring Equipment & 95 & 124 \\
 & Premature Discharge With Incomplete Infection Tx & 71 & 97 \\
\addlinespace
\multirow{5}{*}{\shortstack[l]{Clinical Navigation\\Failures}}
 & Care Coordination \& Transitional Care Gaps & 454 & 1108 \\
 & Lack of Early Post-Discharge Follow-Up & 448 & 914 \\
 & Inadequate Patient Education on Warning Signs & 396 & 739 \\
 & Inadequate Home Support \& Functional Assessment & 294 & 496 \\
 & Medication Access \& Adherence Barriers & 281 & 552 \\
\addlinespace
\multirow{3}{*}{\shortstack[l]{Social Determinants\\of Health}}
 & Missed Inpatient Psychiatric \& Addiction Engagement & 191 & 309 \\
 & Unmanaged Substance Use Triggering Decompensation & 97 & 117 \\
 & Barriers to Outpatient Dialysis Adherence & 36 & 83 \\
\midrule
\multirow{7}{*}{\textit{AI-only (not in Lean)}}
 & Suboptimal COPD Management \& Action Plan & 104 & 232 \\
 & Dietary Non-Adherence & 36 & 36 \\
 & Suboptimal Anticoagulation Management & 31 & 48 \\
 & Undiagnosed or Unmanaged Sleep Apnea & 25 & 31 \\
 & Unmanaged Aspiration Risk \& Dysphagia & 23 & 40 \\
 & Inadequate Steroid Taper \& Outpatient Planning & 23 & 26 \\
 & Continued Tobacco Use Without Cessation Intervention & 19 & 22 \\
\bottomrule
\end{tabular}
\end{table}

\section{Theme Tag Validation}
To check the factor-to-theme assignments, a reviewer who was not involved in defining the themes examined 30 factors sampled uniformly at random from each case study, together with the 57 LOS and 94 readmission theme assignments the LLM had made for them.
For each assigned theme, the reviewer marked whether the theme applied to the factor (yes, no, or unsure), and for each factor, whether an obvious theme was missing.
Excluding the 4 LOS and 5 readmission assignments marked unsure, the reviewer confirmed 83\% of the LOS assignments (95\% CI 74--92\%) and 80\% of the readmission assignments (95\% CI 69--90\%), with intervals from a bootstrap that resamples factors, and flagged no missing theme for any sampled factor.

\section{Heart-Failure Subgroup Theme Comparison}

The prior Lean A3 analysis of readmissions focused on heart-failure patients, whereas the final AI pipeline was applied to four CMS diagnosis cohorts (COPD, heart failure, AMI, and pneumonia).
To assess whether themes found only by the LLM reflect this broader case mix rather than discovery within the Lean population, we recomputed the theme distribution within the heart-failure subgroup (282 of the 484 scale-up encounters; Table~\ref{tab:hf_subgroup}).

The Spearman rank correlation between the full-cohort and subgroup theme prevalences is 0.97, and the ten most prevalent themes are identical.
The heart-failure-specific themes become more prevalent, as expected.
Themes proposed only by the LLM that are clinically relevant to heart failure remain present in the subgroup (Dietary Non-Adherence, 12.4\%; Suboptimal Anticoagulation Management, 6.7\%).

\begin{table}[h]
\centering
\footnotesize
\setlength{\tabcolsep}{4pt}
\caption{Exploratory heart-failure subgroup theme comparison for the
readmission case study. Since the prior Lean A3 analysis focused on heart-failure
patients and the AI pipeline covered four CMS diagnosis cohorts, this
analysis recomputes theme prevalence within the
heart-failure subgroup (282 of the 484 scale-up encounters with
tagged factors).
Cells give the number of encounters
contributing at least one factor to the theme, the percentage of the
population, and the prevalence rank within the population (ties share
rank). Themes are grouped by Lean A3 category as in the full theme table.}
\label{tab:hf_subgroup}
\begin{tabular}{>{\raggedright\arraybackslash}p{2.6cm}>{\raggedright\arraybackslash}p{6.2cm}rrrr}
\toprule
 & & \multicolumn{2}{c}{\textbf{All encounters}} & \multicolumn{2}{c}{\textbf{Heart failure}} \\
\cmidrule(lr){3-4}\cmidrule(lr){5-6}
\textbf{Lean Category} & \textbf{AI-Identified Theme} & n (\%) & Rank & n (\%) & Rank \\
\midrule
 \multirow{7}{=}{Inpatient Care Gaps} & Unresolved Clinical Status at Discharge & 378 (78.1) & 4 & 225 (79.8) & 3 \\
  & Insufficient Heart Failure Post-Discharge Monitoring & 253 (52.3) & 7 & 195 (69.1) & 5 \\
  & Suboptimal Diuretic Regimen/Monitoring & 184 (38.0) & 9 & 149 (52.8) & 8 \\
  & Delayed or Suboptimal GDMT Optimization & 146 (30.2) & 10 & 96 (34.0) & 10 \\
  & Suboptimal HFrEF Medical Therapy \& Monitoring & 113 (23.3) & 11 & 88 (31.2) & 11 \\
  & Lack of Home Monitoring Equipment & 95 (19.6) & 14 & 52 (18.4) & 13 \\
  & Premature Discharge With Incomplete Infection Tx & 71 (14.7) & 15 & 28 (9.9) & 16 \\
\addlinespace
 \multirow{5}{=}{Clinical Navigation Failures} & Care Coordination \& Transitional Care Gaps & 454 (93.8) & 1 & 266 (94.3) & 1 \\
  & Lack of Early Post-Discharge Follow-Up & 448 (92.6) & 2 & 255 (90.4) & 2 \\
  & Inadequate Patient Education on Warning Signs & 396 (81.8) & 3 & 213 (75.5) & 4 \\
  & Inadequate Home Support \& Functional Assessment & 294 (60.7) & 5 & 176 (62.4) & 7 \\
  & Medication Access \& Adherence Barriers & 281 (58.1) & 6 & 179 (63.5) & 6 \\
\addlinespace
 \multirow{3}{=}{Social Determinants of Health} & Missed Inpatient Psychiatric \& Addiction Engagement & 191 (39.5) & 8 & 120 (42.6) & 9 \\
  & Unmanaged Substance Use Triggering Decompensation & 97 (20.0) & 13 & 71 (25.2) & 12 \\
  & Barriers to Outpatient Dialysis Adherence & 36 (7.4) & 16 & 33 (11.7) & 15 \\
\midrule
 \multirow{7}{=}{\textit{AI-only (not in Lean)}} & Suboptimal COPD Management \& Action Plan & 104 (21.5) & 12 & 13 (4.6) & 18 \\
  & Dietary Non-Adherence & 36 (7.4) & 16 & 35 (12.4) & 14 \\
  & Suboptimal Anticoagulation Management & 31 (6.4) & 18 & 19 (6.7) & 17 \\
  & Undiagnosed or Unmanaged Sleep Apnea & 25 (5.2) & 19 & 10 (3.5) & 19 \\
  & Unmanaged Aspiration Risk \& Dysphagia & 23 (4.8) & 20 & 2 (0.7) & 20 \\
  & Inadequate Steroid Taper \& Outpatient Planning & 23 (4.8) & 20 & 1 (0.4) & 21 \\
  & Continued Tobacco Use Without Cessation Intervention & 19 (3.9) & 22 & 1 (0.4) & 21 \\
\bottomrule
\end{tabular}
\end{table}

\section{Cohort Characteristics and Study Period}
\label{sec:cohort_chars}

Tables~\ref{tab:supp_study_period}, \ref{tab:supp_cohort_los}, and~\ref{tab:supp_cohort_readm} report the admission windows and patient characteristics of the encounters used at each stage of both case studies (refinement, evaluation, and scale-up).
In the readmission study, each encounter is a readmission: the admission dates and stay characteristics in Table~\ref{tab:supp_cohort_readm} describe the readmission stay itself, whereas the CMS diagnosis cohort is defined by the index admission that preceded it; refinement encounters drawn before the study narrowed to the four CMS cohorts appear as Not classified.
Cells with one to four patients are shown as $<$5.

\begin{table}[htbp]
\centering
\small
\caption{Calendar window of the encounters used at each stage. Admissions are dated by the encounter under review: the index admission for the length-of-stay study, the readmission itself for the readmission study.}
\label{tab:supp_study_period}
\begin{tabular}{lcc}
\toprule
 & Earliest admission & Latest admission \\
\midrule
\multicolumn{3}{l}{\textbf{Length of stay}} \\
\quad Refinement & Jul 2021 & Oct 2023 \\
\quad Evaluation & Jul 2021 & Dec 2023 \\
\quad Scale-up & Jun 2021 & Feb 2024 \\
\multicolumn{3}{l}{\textbf{Readmission}} \\
\quad Refinement & Mar 2020 & Feb 2023 \\
\quad Evaluation & Dec 2019 & Feb 2024 \\
\quad Scale-up & Aug 2019 & Feb 2024 \\
\bottomrule
\end{tabular}
\end{table}
\begin{table}[htbp]
\centering
\small
\caption{Length of stay study: encounters used at each stage. Median [IQR] or n (\%); counts of one to four are suppressed as $<$5. Diagnosis groups are the ICD-10 parent codes defining the cohort.}
\label{tab:supp_cohort_los}
\begin{tabular}{lccc}
\toprule
 & Refinement & Evaluation & Scale-up \\
\midrule
N encounters & 19 & 51 & 478 \\
Unique patients & 19 & 51 & 465 \\
Age at admission, years & 68.9 [59.8, 83.0] & 61.6 [44.2, 74.7] & 60.7 [45.7, 73.1] \\
\multicolumn{4}{l}{\textit{Sex}} \\
\quad Female & 11 (58\%) & 14 (27\%) & 150 (31\%) \\
\quad Male & 8 (42\%) & 37 (73\%) & 327 (68\%) \\
\quad Another or unknown & 0 (0\%) & 0 (0\%) & $<$5 \\
\multicolumn{4}{l}{\textit{Preferred language}} \\
\quad English & 14 (74\%) & 40 (78\%) & 345 (72\%) \\
\quad Spanish & $<$5 & 5 (10\%) & 66 (14\%) \\
\quad Chinese & $<$5 & $<$5 & 37 (8\%) \\
\quad Other or unknown & $<$5 & $<$5 & 30 (6\%) \\
\multicolumn{4}{l}{\textit{Race}} \\
\quad White & 6 (32\%) & 21 (41\%) & 152 (32\%) \\
\quad Black or African American & $<$5 & 8 (16\%) & 89 (19\%) \\
\quad Asian & $<$5 & 8 (16\%) & 86 (18\%) \\
\quad Native Hawaiian or Other Pacific Islander & 0 (0\%) & 0 (0\%) & 5 (1\%) \\
\quad American Indian or Alaska Native & 0 (0\%) & $<$5 & $<$5 \\
\quad Other & $<$5 & 11 (22\%) & 109 (23\%) \\
\quad More than one race & 5 (26\%) & $<$5 & 34 (7\%) \\
\quad Unknown or declined & 0 (0\%) & 0 (0\%) & $<$5 \\
\multicolumn{4}{l}{\textit{Ethnicity}} \\
\quad Hispanic, Latino/a, or Spanish origin & $<$5 & 8 (16\%) & 111 (23\%) \\
\multicolumn{4}{l}{\textit{Diagnosis group}} \\
\quad Sepsis (A41) & 5 (26\%) & 15 (29\%) & 122 (26\%) \\
\quad Ischemic stroke (I63) & $<$5 & 7 (14\%) & 106 (22\%) \\
\quad Intracranial injury (S06) & 5 (26\%) & 12 (24\%) & 79 (17\%) \\
\quad Drug poisoning (T40) & $<$5 & 8 (16\%) & 75 (16\%) \\
\quad Cellulitis (L03) & $<$5 & 9 (18\%) & 95 (20\%) \\
\quad Not classified & 0 (0\%) & 0 (0\%) & $<$5 \\
Length of stay, days & 5.8 [4.3, 7.2] & 5.7 [4.4, 8.2] & 6.4 [4.7, 9.9] \\
Notes per encounter & 92 [58, 127] & 72 [54, 106] & 81 [59, 126] \\
\bottomrule
\end{tabular}
\end{table}
\begin{table}[htbp]
\centering
\small
\caption{Readmission study: encounters used at each stage. Median [IQR] or n (\%); counts of one to four are suppressed as $<$5. Diagnosis groups are the CMS cohort of the index admission preceding each readmission.}
\label{tab:supp_cohort_readm}
\begin{tabular}{lccc}
\toprule
 & Refinement & Evaluation & Scale-up \\
\midrule
N encounters & 10 & 44 & 484 \\
Unique patients & 10 & 42 & 347 \\
Age at admission, years & 62.6 [44.3, 73.8] & 61.2 [50.5, 69.5] & 62.4 [52.2, 68.7] \\
\multicolumn{4}{l}{\textit{Sex}} \\
\quad Female & $<$5 & 14 (32\%) & 127 (26\%) \\
\quad Male & 6 (60\%) & 30 (68\%) & 356 (74\%) \\
\quad Another or unknown & 0 (0\%) & 0 (0\%) & $<$5 \\
\multicolumn{4}{l}{\textit{Preferred language}} \\
\quad English & 6 (60\%) & 32 (73\%) & 379 (78\%) \\
\quad Spanish & $<$5 & 10 (23\%) & 57 (12\%) \\
\quad Chinese & 0 (0\%) & 0 (0\%) & 20 (4\%) \\
\quad Other or unknown & $<$5 & $<$5 & 28 (6\%) \\
\multicolumn{4}{l}{\textit{Race}} \\
\quad White & $<$5 & $<$5 & 74 (15\%) \\
\quad Black or African American & $<$5 & 19 (43\%) & 195 (40\%) \\
\quad Asian & $<$5 & $<$5 & 68 (14\%) \\
\quad Native Hawaiian or Other Pacific Islander & $<$5 & $<$5 & 12 (2\%) \\
\quad American Indian or Alaska Native & 0 (0\%) & 0 (0\%) & $<$5 \\
\quad Other & $<$5 & 13 (30\%) & 82 (17\%) \\
\quad More than one race & $<$5 & $<$5 & 52 (11\%) \\
\quad Unknown or declined & 0 (0\%) & 0 (0\%) & 0 (0\%) \\
\multicolumn{4}{l}{\textit{Ethnicity}} \\
\quad Hispanic, Latino/a, or Spanish origin & $<$5 & 13 (30\%) & 87 (18\%) \\
\multicolumn{4}{l}{\textit{Diagnosis group}} \\
\quad Heart Failure & $<$5 & 25 (57\%) & 282 (58\%) \\
\quad COPD & $<$5 & 7 (16\%) & 113 (23\%) \\
\quad Pneumonia & $<$5 & 5 (11\%) & 52 (11\%) \\
\quad AMI & $<$5 & 7 (16\%) & 36 (7\%) \\
\quad Stroke & 0 (0\%) & 0 (0\%) & 0 (0\%) \\
\quad Not classified & $<$5 & 0 (0\%) & $<$5 \\
Length of stay, days & 5.4 [3.3, 6.7] & 3.7 [1.8, 5.8] & 4.1 [2.1, 7.5] \\
Notes per encounter & 65 [34, 89] & 40 [24, 68] & 42 [23, 72] \\
\bottomrule
\end{tabular}
\end{table}
 
\section{Generic-Prompt (Round 1) Theme Coverage}
\label{sec:v1_ablation}

Here we present a comparison of factors extracted by the first (v1) prompt versus the final prompt in the LOS case study.
Because the v1 prompt was constructed with minimal clinical input, this represents a generic prompt and was constructed completely independently of any prior Lean analysis results; the v1 round also used a different extraction model (GPT-5 Mini) than the final pipeline.
We tagged 179 v1 factors against the final 27 LOS themes with the same tagging model and prompt, without re-clustering.
The v1 factors covered 26 of the 27 final themes, and only the theme on prolonged emergency department boarding was absent.
Because the themes were defined by the final pipeline, this analysis shows that the generic prompt's factors cover the final taxonomy, not that the generic prompt independently recovers or validates these themes.

Nevertheless, there are two key differences between the v1 and final LOS analyses.
First, the tuned pipeline produces QI factors that are generally more clinically specific: for example, for the same encounter, the v1 prompt surfaced that inpatient methadone titration and coordination with the outpatient clinic prolonged the stay, whereas the final pipeline identified that the patient had already been titrated to a therapeutic dose during an admission one month earlier, but a failed handoff to outpatient treatment forced a complete re-titration and four additional inpatient days.
Second, the rankings between the themes across v1 and the final prompts are aligned but still differ appreciably: the Spearman rank correlation between the v1 and published theme prevalences is 0.57.
Of note, themes that originally received low rankings in the v1 outputs (e.g., weekend service limitations, echocardiogram delays, addiction care team consultation) increased in ranking through the co-optimization process, which clarified the modifiability and role of these factors in the specific clinical setting.
The different extraction model and largely different encounters in the v1 round may also contribute to these changes in ranking.

\section{Prior Manual Lean Analyses}

The AI pipeline results were compared against manual Lean analyses previously conducted at ZSFG for both hospital metrics.

\subsection{LOS}

The LOS Lean analysis was part of ZSFG's Executive Strategy A3 for optimizing patient access and flow. The effort spanned roughly six months. The QI team reviewed approximately 25 encounters across the five target diagnosis groups, collecting all inpatient notes for each encounter. These notes were summarized using an LLM, and clinical experts synthesized the summaries into ``prototypical scenarios''---composite patient journeys representing the typical care trajectory for each DRG. Frontline clinicians then mapped these prototypical journeys using Value Stream Mapping to identify bottlenecks and delays at each stage. The analysis produced six high-level categories of modifiable flow barriers: Discharge Planning \& Post-Acute Coordination, Role Clarity (SW, UM, ACT), Capacity \& Coverage, Consult Timeliness \& Coordination, Diagnostic/Procedural Workflow \& Turnaround, and Resource Utilization \& Capacity Management.

\subsection{Readmission}

The readmission Lean analysis followed the A3 problem-solving methodology and focused on reducing 30-day all-cause readmissions among patients with an index heart-failure admission. The A3 documented the current state of heart-failure readmissions at ZSFG, set measurable reduction targets, and identified top contributing factors through chart review and clinical team input. Contributors were organized into three categories: (1) inpatient care gaps (e.g., patients discharged before achieving euvolemia, no follow-up within 7 days, low referral rates to advanced heart-failure services or hospice), (2) clinical navigation failures (e.g., inability to obtain medications, unawareness of follow-up appointments, minimal coordination between inpatient, outpatient specialty, mental health/substance use, and primary care providers), and (3) social determinants of health (e.g., homelessness, low health literacy, stimulant or alcohol use disorders).

\section{Learning Process Details for Case Study 1: Length of Stay}

\textbf{Round 1 $\rightarrow$ 2: Separating reasoning from evidence for traceability.}
 The initial prompt design (v1) established a two-step value stream mapping approach: first creating a Gantt chart mapping the patient journey, then extracting LOS factors with a single ``explanation'' field combining reasoning and evidence.
 However, the team found it difficult to verify whether the LLM's reasoning was actually supported by the clinical notes.
 For Round 2, the team separated the single \texttt{explanation} field into two distinct fields: \texttt{Explanation} for detailed step-by-step reasoning and \texttt{Evidence} for exact quotes supporting each reasoning step.
 This improved traceability between the LLM's conclusions and the source documentation.

 \textbf{Round 2 $\rightarrow$ 4: Calibration and distinguishing illness severity from process delays.}
 Round 3 introduced probability ranges to the confidence scale (e.g., 3=definitely true at $>$95\%, 2=likely true at 70--95\%) to enable more precise calibration.
 However, expert review revealed a critical issue: the LLM was conflating patient acuity with modifiable process factors.
 For instance, the LLM would flag ``patient required prolonged ICU stay'' as an LOS factor, when this was due to illness severity rather than operational inefficiency.
 Round 4 addressed this by adding explicit instructions to ``differentiate between medical events that were unavoidable due to illness severity and structural delays within the process of care coordination, patient flow and discharge.''
 The prompt also added a benchmark: ``Expected time for imaging procedures and consult completion from order placement to completion is approximately 24 hours. Imaging completion beyond this timeframe should be considered intervenable.''

 \textbf{Round 4 $\rightarrow$ 6: Actionability and critical thinking.}
 Round 5 added a \texttt{process\_improvement} field requiring the LLM to propose specific interventions for each identified factor.
 Critically, the prompt now included explicit lists of NON-actionable observations (e.g., ``No ICU/step-down beds available,'' ``SNF or rehab facilities at capacity,'' ``Insurance denied authorization'') versus ACTIONABLE process improvements (e.g., ``Late initiation of discharge planning,'' ``Late bed requests,'' ``Equipment not ordered early'').
 Round 6 further refined the approach by splitting the explanation into \texttt{explanation\_support} (why the factor lengthened LOS) and \texttt{explanation\_contrary} (why it may NOT have lengthened LOS).
 Requiring contrary evidence improved critical thinking and reduced over-attribution of LOS to spurious factors.

 \textbf{Round 6 $\rightarrow$ Final: Architecture, calibration, and framing alignment.}
 Round 7 introduced two major changes.
 First, an architecture change: confidence scoring was split into a separate LLM call, reducing output complexity per call and allowing independent tuning of confidence calibration.
 Second, the confidence scale changed from a Likert scale (1--3) to a percentage scale (0--100, rounded to the nearest decile), with ZSFG clinician-annotated examples anchoring expectations (e.g., ``Weekend PT service limitations $\rightarrow$ 90\% confidence'').
 Round 8 reframed the entire analysis from ``operational factors'' to ``modifiable gaps,'' with language emphasizing ``patient flow'' rather than just LOS reduction.
 ``Patient flow'' was used because this was what the humans were truly assessing for, rather than overall LOS reduction -- this was noticed because the LLM would try to reason about how different factors/events interacted and whether any events would mask the effects of another factor, whereas the human experts evaluated each factor individually, which the term `patient flow' captured most accurately.
 The prompt also clarified that ``Bed capacity cannot be increased directly. Instead the hospital will address other modifiable factors that will indirectly increase bed capacity,'' because the LLM would commonly point out bed capacity as a reason for why LOS was longer but the hospital QI team decided this was not a directly modifiable factor; rather, this was more often an indirectly modifiable factor, assuming the team could make processes more efficient at the hospital.

\section{Learning Process Details for Case Study 2: Readmission}

\textbf{Round 1 $\rightarrow$ 3: Optimizing clinical context and emphasizing causal chains.}
The initial prompt revealed that the LLM often identified potentially relevant factors but lacked sufficient clinical context to make confident assessments.
For example, in one case involving an incomplete goals of care discussion, the LLM noted relevant quotes about the family ``moving toward DNR/DNI but needs time to discuss as a group,'' but the available documentation was insufficient to determine whether earlier intervention would have prevented readmission.
Based on expert feedback, Round 2 expanded the clinical context by adding consult notes, outpatient notes, and care plan notes.
However, expert review found that care plan notes were mostly templated nursing boilerplate (e.g., ``Goal: verbalizes/displays adequate comfort level. Outcome: Progressing'') and added noise rather than signal.
Round 3 removed care plan notes while retaining the more informative consult notes, and updated the prompt to emphasize causal chains more explicitly---requiring the LLM to trace a clear path from the index admission gap to the readmission event.

\textbf{Round 3 $\rightarrow$ 5: Modifiability, realistic interventions, and population refinement.}
Expert annotation revealed several issues with the LLM's reasoning.
First, the LLM sometimes conflated factors that {\em contributed} to readmission with factors that were {\em modifiable}.
For example, discharge while volume-overloaded ``contributed but was not modifiable'' when ``the patient self-directed their discharge.''
Second, the LLM proposed oversimplified interventions that experts found unrealistic: ``if all that was needed to get patients to adhere is a pharmacy phone call it would be an easy fix.''
Third, the LLM made attribution errors, blaming the system when patient factors were primary (e.g., ``Appointment was scheduled, patient didn't show'' or ``They attempted to reach the patient but he changed his phone number'').
Round 4 addressed these issues by reinstating the modifiability requirement and providing explicit examples of realistic process changes.
Round 5 narrowed the patient population to CMS readmission diagnosis groups (COPD, HF, AMI, Pneumonia), where clinical experts had moderate expertise and cases were less idiosyncratic.

\textbf{Round 5 $\rightarrow$ 6: Root cause analysis and architecture optimization.}
Dual expert annotation in Round 5 revealed that the LLM often focused on proximate causes rather than root causes.
For one heart-failure case, an expert noted ``more comes down to social barriers than diuretic dosing''---the diuretic dosing issue was a symptom, not the root cause.
The LLM also sometimes undervalued cases where processes existed but were not followed (e.g., case management enrollment not completed, blood cultures not awaited), which experts viewed as more actionable than missing processes.
Round 6 updated the prompt to emphasize root cause identification and request more detailed process improvement suggestions.
Round 6 also introduced an architecture change: the complex extraction was split into two separate LLM calls (Gantt chart extraction, then factor analysis) to address timeout errors caused by the nested output schema with validators.
This split reduced output complexity per call and improved pipeline reliability.

\section{Full Specifications}
\label{sec:full_specs}

This section documents the complete high-level specifications at each iteration of the Human-AI Spec-Solution Co-Optimization process for both case studies.
Each spec sheet records what the team decided for every component of the AI/ML development pipeline (mapped to AI/ML development stages in Table~1 of the main text) at that iteration.
For the first and final iterations, we provide a complete specification; for intermediate iterations, we document only the components that changed, with all other specifications carried forward from the previous round.
Plain-language descriptions are followed by key verbatim language from the LLM prompts in quotes where applicable.
The full prompt text for all versions is provided in Supplementary Section~\ref{sec:prompt_specs}.
Throughout, the study data scientist authored all prompts, and the six clinical reviewers are denoted Reviewer~1--6; the numbering is consistent across rounds and case studies and does not correspond to author order.

\subsection{Case Study 1: Length of Stay}

\subsubsection{Initial Specifications}

\begin{description}
\item[1. Problem Formalization]\mbox{}
  \begin{description}
  \item[Objective] Identify operational factors that would shorten hospital length of stay: ``contributing operational factors that, if modified, would likely shorten hospital stay.''
  \item[Population] Adult inpatients in the top five DRGs at ZSFG (Sepsis, Skin and Soft Tissue Infection, Ischemic Stroke, Blunt Head Injury, Drug Poisoning) with LOS between 4 and 20 days.
  \item[Label definition] 4-point Likert scale (0--3): 0 = unlikely, 1 = possibly true (implied in note), 2 = likely true (heavily implied), 3 = definitely true (stated in note). A single score captured both causality and modifiability.
  \end{description}

\item[2. Model Learning]\mbox{}
  \begin{description}
  \item[Estimator inputs] All clinical notes and order events for the inpatient encounter, concatenated into a single text block.
  \item[Estimator output] Single JSON object: (1) Gantt chart with timestamped events, (2) list of contributing factors, each with a combined \texttt{explanation} field, \texttt{relevant\_quotes}, and \texttt{confidence} (0--3 Likert).
  \item[Model family] GPT-5 Mini (\texttt{gpt-5-mini-2025-08-07}).
  \item[Prompt tuning] Single LLM call producing both Gantt chart and factors. Two-step structure: Step 1 creates the Gantt chart, Step 2 identifies LOS factors. No in-context learning examples with clinician-annotated scores. Temperature 1.
  \end{description}

\item[3. Model Validation]\mbox{}
  \begin{description}
  \item[What gets validated] Gantt chart, extracted factors, explanation, quotes, confidence scores.
  \item[How output is validated] Low-cost single reader (data scientist), 2 patients.
  \end{description}
\end{description}

\subsubsection*{Iteration 2 (v2)}

\noindent\textit{Prompt author:} data scientist. \textit{Reviewed by:} data scientist (5 patients). \textit{LLM:} GPT-5 Mini.

\noindent All specifications unchanged from v1 except:

\begin{description}
\item[2. Model Learning]\mbox{}
  \begin{description}
  \item[Estimator output] Separated the single \texttt{explanation} field into \texttt{Explanation} (step-by-step reasoning) and \texttt{Evidence} (exact quotes that ``support every step in your step-by-step reasoning'').
  \end{description}

\item[3. Model Validation]\mbox{}
  \begin{description}
  \item[How output is validated] Low-cost single reader (data scientist), 5 patients.
  \end{description}
\end{description}

\subsubsection*{Iteration 3 (v3)}

\noindent\textit{Prompt author:} data scientist. \textit{Reviewed by:} data scientist (10 patients). \textit{LLM:} GPT-5 Mini.

\noindent All specifications unchanged from v2 except:

\begin{description}
\item[1. Problem Formalization]\mbox{}
  \begin{description}
  \item[Objective] Narrowed to ``contributing factors that are actionable and possibly increased LOS by at least 12 hours.''
  \item[Label definition] Added probability ranges: 3 = definitely true ($>$95\%), 2 = likely true (70--95\%), 1 = possibly true (50--70\%), 0 = unlikely ($<$50\%).
  \end{description}

\item[3. Model Validation]\mbox{}
  \begin{description}
  \item[How output is validated] Low-cost single reader (data scientist), 10 patients.
  \end{description}
\end{description}

\subsubsection*{Iteration 4 (v4)}

\noindent\textit{Prompt author:} data scientist. \textit{Reviewed by:} data scientist and Reviewer 1 dual review (10 patients). \textit{LLM:} GPT-5.

\noindent All specifications unchanged from v3 except:

\begin{description}
\item[2. Model Learning]\mbox{}
  \begin{description}
  \item[Model family] GPT-5 (\texttt{gpt-5-2025-08-07}).
  \item[Prompt tuning] Added instruction to ``differentiate between medical events that were unavoidable due to illness severity and structural delays within the process of care coordination, patient flow and discharge.'' Added: ``Do not report factors that extend length of stay because of a patient's illness severity.'' Added 24-hour benchmark for imaging/consult completion.
  \end{description}

\item[3. Model Validation]\mbox{}
  \begin{description}
  \item[How output is validated] Medium-cost single reader (clinician), 10 patients.
  \end{description}
\end{description}

\subsubsection*{Iteration 5 (v5)}

\noindent\textit{Prompt author:} data scientist. \textit{Reviewed by:} Reviewer 1 (10 patients). \textit{LLM:} Claude Opus 4.5.

\noindent All specifications unchanged from v4 except:

\begin{description}
\item[2. Model Learning]\mbox{}
  \begin{description}
  \item[Estimator output] Added \texttt{process\_improvement} field for each factor.
  \item[Model family] Claude Opus 4.5 (\texttt{us.anthropic.claude-opus-4-5-20251101-v1:0}).
  \item[Prompt tuning] Added explicit lists of NON-actionable observations (e.g., ``No ICU/step-down beds available,'' ``SNF or rehab facilities at capacity,'' ``Insurance denied authorization'') and ACTIONABLE process improvements (e.g., ``Late initiation of discharge planning,'' ``Late bed requests,'' ``Lack of weekend hospital services'').
  \end{description}

\item[3. Model Validation]\mbox{}
  \begin{description}
  \item[What gets validated] Added process improvement suggestions.
  \item[How output is validated] Low-cost single reader (clinician), 10 patients.
  \end{description}
\end{description}

\subsubsection*{Iteration 6 (v6)}

\noindent\textit{Prompt author:} data scientist. \textit{Reviewed by:} Reviewer 1 (5 patients, hardcoded to specific encounters for comparison to prior reviews). \textit{LLM:} Claude Opus 4.5.

\noindent All specifications unchanged from v5 except:

\begin{description}
\item[1. Problem Formalization]\mbox{}
  \begin{description}
  \item[Objective] Confidence question changed to counterfactual framing: ``If we assume proposed process changes are implemented for ALL listed reasons EXCEPT for this one, then this reason prolongs the LOS by 6+ hours.''
  \item[Label definition] Simplified to 3-point scale (1--3): 3 = very likely ($>$90\%), 2 = likely (70--90\%), 1 = possibly true (50--70\%).
  \end{description}

\item[2. Model Learning]\mbox{}
  \begin{description}
  \item[Estimator output] Split \texttt{explanation} into \texttt{explanation\_support} and \texttt{explanation\_contrary}. Capped at 5 factors per patient.
  \end{description}

\item[3. Model Validation]\mbox{}
  \begin{description}
  \item[What gets validated] Added supportive and contrary evidence.
  \item[How output is validated] Medium-cost single reader (clinician), 5 patients.
  \end{description}
\end{description}

\subsubsection*{Iteration 7 (v7)}

\noindent\textit{Prompt author:} data scientist. \textit{Reviewed by:} Group kickoff (Reviewers 1--6) --- 52 patients, 4 reviewed by all annotators. \textit{LLM:} Claude Opus 4.5.

\noindent All specifications unchanged from v6 except:

\begin{description}
\item[1. Problem Formalization]\mbox{}
  \begin{description}
  \item[Objective] Confidence statement changed to: ``This contributing factor represents an excessive delay or suboptimal hospital process that ZSFG can further optimize through process changes.'' Counterfactual framing dropped; factor identification threshold reverted to 12 hours.
  \item[Label definition] Replaced 3-point Likert with continuous 0--100\% probability, rounded to nearest decile.
  \end{description}

\item[2. Model Learning]\mbox{}
  \begin{description}
  \item[Prompt tuning] Split into two-stage pipeline: (1) extraction call for Gantt chart and factors, (2) separate scoring call for confidence. LLM role set to ``an experienced clinical quality improvement researcher at ZSFG.'' Added three ZSFG clinician-annotated ICL examples with confidence scores (Weekend PT $\rightarrow$ 90\%, Late SNF coordination $\rightarrow$ 80\%, Delayed TTE $\rightarrow$ 90\%). Actionable factors list restructured as ``HIGHLY ACTIONABLE ($\ge$90\%).''
  \end{description}

\item[3. Model Validation]\mbox{}
  \begin{description}
  \item[How output is validated] High-cost multi-reader (six clinical experts), 52 patients. 4 patients reviewed by all annotators.
  \end{description}
\end{description}

\subsubsection{Final Specifications}

\noindent\textit{Prompt author:} data scientist. \textit{Reviewed by:} Group review (Reviewers 1--6) --- 52 patients, 4 reviewed by all annotators. \textit{LLM:} Claude Opus 4.5.

\begin{description}
\item[1. Problem Formalization]\mbox{}
  \begin{description}
  \item[Objective] ``This contributing factor is a modifiable gap that if improved would streamline patient flow.'' Bed capacity explicitly excluded as a directly modifiable factor.
  \item[Population] Adult inpatients in the top five DRGs at ZSFG with LOS between 4 and 20 days. (Unchanged from v1.)
  \item[Label definition] Expert annotation: 1--5 Likert scale. AI output: 0--100\% probability, rounded to nearest decile, calibrated against the Likert scale post hoc.
  \end{description}

\item[2. Model Learning]\mbox{}
  \begin{description}
  \item[Estimator inputs] All clinical notes and order events for the inpatient encounter. (Unchanged from v1.)
  \item[Estimator output] Three-stage: (1) Gantt chart JSON with timestamped events; (2) per factor: \texttt{reason}, \texttt{explanation\_support}, \texttt{explanation\_contrary}, \texttt{relevant\_quotes}, \texttt{process\_improvement}; (3) \texttt{confidence} (0--100\%) in separate LLM call. Three clinician-annotated ICL examples. Explicit lists of highly actionable ($\ge$90\%) and non-actionable factors. At most 5 factors, each must have increased LOS by 12+ hours.
  \item[Model family] Claude Opus 4.5 (\texttt{us.anthropic.claude-opus-4-5-20251101-v1:0}).
  \item[Prompt tuning] Three-stage pipeline: (1) Gantt chart extraction, (2) factor extraction given Gantt chart, (3) confidence scoring given Gantt chart and factors. Factors must be both modifiable AND causal. Illness severity explicitly excluded. 24-hour benchmark for consult/imaging completion.
  \end{description}

\item[3. Model Validation]\mbox{}
  \begin{description}
  \item[What gets validated] Gantt chart, factors, supportive/contrary reasoning, quotes, process improvements, confidence scores.
  \item[How output is validated] High-cost multi-reader (six assigned clinical experts). 1--5 Likert scale via validation UI. 52 assigned patients; 4 reviewed by all annotators. Inter-rater agreement on the factors rated by all reviewers, exact: 31.5\%, within-one-point: 72.6\%. LLM-rater agreement on all ratings, exact: 32.1\%, within-one-point: 76.7\%.
  \end{description}
\end{description}

\subsection{Case Study 2: 30-Day Unplanned Readmission}

This case study was conducted after the LOS case study, so the initial specifications carried over several design decisions (Claude Opus 4.5, 0--100\% confidence scale, supportive/contrary evidence fields, separate scoring prompt).

\subsubsection{Initial Specifications}

\begin{description}
\item[1. Problem Formalization]\mbox{}
  \begin{description}
  \item[Objective] ``This contributing factor is a modifiable gap that if improved would likely have prevented this readmission.'' Factors must be both modifiable and causal.
  \item[Population] All adult patients with 30-day unplanned readmissions at ZSFG. No diagnosis group filtering.
  \item[Label definition] 0--100\% probability, rounded to nearest decile.
  \end{description}

\item[2. Model Learning]\mbox{}
  \begin{description}
  \item[Estimator inputs] Index admission: admission notes. Readmission: ED note, admission note, H\&P. No outpatient, consult, or discharge instruction notes.
  \item[Estimator output] Two-stage: (1) Gantt chart + factor extraction in one call (with \texttt{readmission\_summary}), each factor including \texttt{reason}, \texttt{explanation\_support}, \texttt{explanation\_contrary}, \texttt{relevant\_quotes}, 
  
  \noindent \texttt{process\_improvement}, \texttt{confidence}; (2) separate confidence scoring. Three ICL examples. Lists of actionable (9 items) and non-actionable (6 items) factors. At most 5 factors; minimum 50\% confidence.
  \item[Model family] Claude Opus 4.5 (\texttt{us.anthropic.claude-opus-4-5-20251101-v1:0}).
  \item[Prompt tuning] Two-stage pipeline: (1) combined Gantt chart and factor extraction, (2) separate confidence scoring.
  \end{description}

\item[3. Model Validation]\mbox{}
  \begin{description}
  \item[What gets validated] Gantt chart, factors, supportive/contrary reasoning, quotes, process improvements, confidence scores.
  \item[How output is validated] Low-cost multi-reader (data scientist + clinicians), 4 patients. Only first reason reviewed per patient.
  \end{description}
\end{description}

\subsubsection*{Iteration 2 (v2)}

\noindent\textit{Prompt author:} data scientist. \textit{Reviewed by:} Reviewer 5 (4 patients). \textit{LLM:} Claude Opus 4.5.

\noindent Extraction prompt unchanged from v1. Change was in data assembly:

\begin{description}
\item[2. Model Learning]\mbox{}
  \begin{description}
  \item[Estimator inputs] Added consult notes, outpatient notes, and care plan notes.
  \end{description}

\item[3. Model Validation]\mbox{}
  \begin{description}
  \item[How output is validated] Medium-cost single reader (clinician), 4 patients.
  \end{description}
\end{description}

\subsubsection*{Iteration 3 (v3)}

\noindent\textit{Prompt author:} data scientist. \textit{Reviewed by:} data scientist (4 patients). \textit{LLM:} Claude Opus 4.5.

\noindent All specifications unchanged from v2 except:

\begin{description}
\item[2. Model Learning]\mbox{}
  \begin{description}
  \item[Estimator inputs] Removed care plan notes. Added discharge instructions from index admission. Retained consult and outpatient notes.
  \item[Prompt tuning] Shifted to root cause analysis framing: ``identify the specific factors that likely caused or contributed to this readmission.'' Added causal chain questions: ``What specific decision, action, or omission occurred? How did this directly lead to the clinical state that required readmission?'' Restructured actionable factor categories into three headers: Clinical Management, Care Coordination, Patient Education. Updated ICL examples to show causal chains.
  \end{description}

\item[3. Model Validation]\mbox{}
  \begin{description}
  \item[How output is validated] Low-cost single reader (data scientist), 4 patients.
  \end{description}
\end{description}

\subsubsection*{Iteration 4 (v4)}

\noindent\textit{Prompt author:} data scientist. \textit{Reviewed by:} Reviewer 1 (4 patients). \textit{LLM:} Claude Opus 4.5.

\noindent All specifications unchanged from v3 except:

\begin{description}
\item[2. Model Learning]\mbox{}
  \begin{description}
  \item[Estimator inputs] Removed consult notes from the readmission encounter.
  \item[Prompt tuning] Added explicit examples of factors that are NOT modifiable (e.g., ``patient self-directed their discharge,'' ``Appointment was scheduled, patient didn't show''). Added ``Provide evidence-based recommendations'' to process improvement instructions. Scope expanded to ``ZSFG or surrounding hospitals.''
  \end{description}

\item[3. Model Validation]\mbox{}
  \begin{description}
  \item[How output is validated] Medium-cost single reader (clinician), 4 patients.
  \end{description}
\end{description}

\subsubsection*{Iteration 5 (v5)}

\noindent\textit{Prompt author:} data scientist. \textit{Reviewed by:} data scientist and Reviewer 1 dual review (8 patients, all reviewed by both). \textit{LLM:} Claude Opus 4.5.

\noindent All specifications unchanged from v4 except:

\begin{description}
\item[1. Problem Formalization]\mbox{}
  \begin{description}
  \item[Population] Narrowed to CMS readmission diagnosis groups: COPD, Heart Failure, AMI, and Pneumonia.
  \end{description}

\item[3. Model Validation]\mbox{}
  \begin{description}
  \item[How output is validated] Medium-cost multi-reader (two clinicians), 8 patients with dual annotation.
  \end{description}
\end{description}

\subsubsection{Final Specifications}

\noindent\textit{Prompt author:} data scientist. \textit{Reviewed by:} Group review (five of Reviewers 1--6) --- 44 patients, 4 reviewed by all annotators. \textit{LLM:} Claude Opus 4.5.

\begin{description}
\item[1. Problem Formalization]\mbox{}
  \begin{description}
  \item[Objective] ``This contributing factor is a modifiable gap that if improved would reduce readmission risk.'' The AND requirement made explicit: ``we are looking for factors that are both modifiable AND causal.''
  \item[Population] CMS readmission diagnosis groups: COPD, Heart Failure, AMI, Pneumonia. (Unchanged from v5.)
  \item[Label definition] Expert annotation: 1--5 Likert scale. AI output: 0--100\% probability, rounded to nearest decile.
  \end{description}

\item[2. Model Learning]\mbox{}
  \begin{description}
  \item[Estimator inputs] Index admission: consult notes, discharge summary, discharge instructions. Intervening outpatient notes. Readmission: ED provider note, H\&P, discharge summary. Excluded: care plan notes, readmission consult notes.
  \item[Estimator output] Three-stage: (1) Gantt chart spanning index admission through readmission; (2) per factor: \texttt{reason}, \texttt{explanation\_support}, \texttt{explanation\_contrary}, \texttt{relevant\_quotes}, \texttt{process\_improvement} (confidence removed from extraction, delegated to scoring stage); (3) \texttt{confidence} (0--100\%) in separate scoring call. Three ICL examples with causal chains. Lists of high-confidence actionable factors (Clinical Management, Post-Discharge Care Coordination, Patient Education) and low-confidence/non-actionable factors.
  \item[Model family] Claude Opus 4.5 (\texttt{us.anthropic.claude-opus-4-5-20251101-v1:0}).
  \item[Prompt tuning] Three-stage pipeline: (1) Gantt chart extraction, (2) factor extraction given Gantt chart, (3) confidence scoring. Split from two to three stages to address API timeout errors. Extraction prompt emphasized root cause over proximate cause. ``Post-Discharge Care Coordination'' category expanded to include modifiable social circumstances.
  \end{description}

\item[3. Model Validation]\mbox{}
  \begin{description}
  \item[What gets validated] Gantt chart, factors, supportive/contrary reasoning, quotes, process improvements, confidence scores.
  \item[How output is validated] High-cost multi-reader (five assigned clinical experts). 1--5 Likert scale via validation UI. 44 assigned patients; 4 reviewed by all annotators. Inter-rater agreement on the factors rated by all reviewers, exact: 23.0\%, within-one-point: 72.5\%. LLM-rater agreement on all ratings, exact: 28.3\%, within-one-point: 71.0\%.
  \end{description}
\end{description}
 
\section{Final Prompts}
\label{sec:prompt_specs}

The final AI pipeline for each case study used a three-stage architecture: Stage~1 extracts a Gantt chart of the patient journey from the clinical notes, Stage~2 identifies modifiable contributing factors given the Gantt chart, and Stage~3 assigns confidence scores to each factor in a separate LLM call. In the prompts below, \texttt{<CLINICAL NOTES INSERTED HERE>} is replaced at runtime with the patient's clinical documentation, \texttt{<GANTT CHART JSON FROM STAGE 1 INSERTED HERE>} with the Stage~1 output, and \texttt{<EXTRACTED FACTORS JSON FROM STAGE 2 INSERTED HERE>} with the Stage~2 output. All prompts instruct the LLM to return structured JSON; the pipeline enforces the output schema programmatically.

\subsection{Length of Stay}

\subsubsection*{Stage 1: Gantt Chart Extraction}
\begin{Verbatim}[breaklines,fontsize=\small]
You are an experienced clinical quality improvement researcher at ZSFG and analyzing inpatient length of stay at your hospital, using a two-step value stream mapping approach. This follows Lean Healthcare methodology to first map the patient journey, then identify contributing factors that are modifiable gaps and if improved would likely shorten hospital stay. As an experienced clinical QI researcher, be confident in making suggestions -- leverage your expertise!

The patient encounter timeline is fully captured below, which includes ALL clinical events and ALL notes written for this patient encounter:
<CLINICAL NOTES INSERTED HERE>

=== STEP 1: VALUE STREAM MAPPING - PATIENT JOURNEY GANTT CHART ===

Create a Gantt chart that maps the key phases of the patient's hospital stay. Focus on essential care phases, treatment periods, and potential bottlenecks/delays that may have contributed to extended length of stay (LOS). Differentiate between medical events that were unavoidable due to illness severity and structural delays within the process of care coordination, patient flow, and discharge.

Guidelines for creating the Gantt chart:
1. **Map the patient journey, emphasizing events that extended LOS**: Capture essential care phases, major treatments, and delays that extended the hospital stay
2. **Include entire hospital timeline**: Cover admission through discharge, noting when the patient was medically ready for discharge vs. actual discharge
3. **Identify bottlenecks**: Note waiting periods, care coordination delays, resource availability issues, etc (if any)
4. **Assign event timings**: Assign event timing. If exact timestamps aren't available, provide reasonable estimates and mark them as approximate. If there are important events that extend beyond discharge, set the end timestamp to the time of discharge.

Event categories to consider, though you can introduce others:
- **admission**: Initial care phases (ED, admission process)
- **treatment**: Active medical treatment periods (IV medications, monitoring phases)
- **procedure**: Specific interventions, surgeries, or procedures
- **waiting**: Delays, waiting for results, scheduling gaps, resource availability
- **coordination**: Care transitions, discharge planning, specialist consultations
- **discharge**: Final discharge processes, waiting for equipment/placement

Output the Gantt chart in the structured JSON format given below. Example output:
{
  "index_admission_summary": "Brief summary of the stated reason for patient's admission. Summarize only, no interpretation.",
  "events": [
    {
      "event_id": 1,
      "label": "IV Antibiotic Treatment",
      "category": "treatment",
      "description": "Intravenous antibiotic course for infection",
      "start_time": "2024-01-10 08:00",
      "end_time": "2024-01-15 12:00",
      "relevant_quotes": "Started on IV vancomycin for surgical site infection"
    },
    {
      "event_id": 2,
      "label": "Waiting for Outpatient Imaging",
      "category": "waiting",
      "description": "Delay for outpatient MRI availability despite medical readiness for discharge",
      "start_time": "2024-01-15 10:00",
      "end_time": "2024-01-18 14:00",
      "time_uncertainty": "estimated",
      "relevant_quotes": "Patient ready for discharge...waiting 3 days for outpatient MRI availability"
    }
  ]
}

Output requirements:
* The `relevant_quotes` field must contain EXACT quotes from the clinical notes - word-for-word, with no paraphrasing. Use "..." to connect quotes from different sections.
\end{Verbatim}

\subsubsection*{Stage 2: Factor Extraction}
\begin{Verbatim}[breaklines,fontsize=\small]
You are an experienced clinical quality improvement researcher at ZSFG and analyzing inpatient length of stay at your hospital, using a two-step value stream mapping approach. This follows Lean Healthcare methodology to first map the patient journey, then identify contributing factors that are modifiable gaps and if improved would likely shorten hospital stay. As an experienced clinical QI researcher, be confident in making suggestions -- leverage your expertise!

The patient encounter timeline is fully captured below, which includes ALL clinical events and ALL notes written for this patient encounter:
<CLINICAL NOTES INSERTED HERE>

=== STEP 1: VALUE STREAM MAPPING - PATIENT JOURNEY GANTT CHART ===

The following Gantt chart has already been created mapping the key phases of the patient's hospital stay:

<GANTT CHART JSON FROM STAGE 1 INSERTED HERE>

=== STEP 2: FACTORS ANALYSIS ===

Go through the Gantt chart above to identify MODIFIABLE factors that likely caused or contributed to extended length of stay. This is an AND statement -- we are looking for factors that are both modifiable AND causal.

Instructions:
1. Going through the Gantt chart, identify opportunities where there was excessive delay, suboptimal coordination/processes, or prolonged duration, which likely led to LOS being lengthened by 12+ hours. Only list contributors that are actionable, such as resource availability, guideline-directed medical therapy, care coordination issues; avoid listing a patient's illness severity as a factor. Prioritize listing factors that very likely increased LOS by 12+ hours. List at most 5 factors.
2. Explanation Support: Provide detailed step-by-step reasoning for why this represents a contributing factor that led to prolonged LOS or suboptimal patient flow, referencing both the Gantt chart timeline and clinical notes when applicable.
3. Explanation Contrary: Provide explanations for why this factor may not need to be or cannot be optimized further.
4. Relevant Quotes: For each identified contributing factor, provide EXACT quotes (word-for-word) from the note. Quotes should support all components of your explanation.
5. Process Improvement: For each factor, describe what specific process change could be implemented, which may ultimately shorten LOS. Focus on timing and workflow changes within the hospital's control.

Example categories of factors to consider:

HIGHLY ACTIONABLE factors (should be assigned high confidence >= 90%
- Lack of weekend hospital services
   => Add weekend hospital service
- Needing better parallelization of tasks, such as initiating discharge planning late (after patient medically stable rather than proactively)
   => Can be optimized by starting discharge planning early or better parallelization of tasks
- Social work consulted only after barriers emerged rather than at admission
   => Early SW involvement should be initiated by the team or proactively
- Insurance pre-authorization not initiated early in admission
   => Initiate insurance pre-authorization as soon as possible
- Care conference scheduled late, delaying family decision-making
   => Family decision-making can begin much earlier, with SW beginning early conversations with family
- Anticipated equipment needs not ordered until day of discharge
   => known equipment needs should be ordered asap, so that it does not delay discharge
- Delayed specialist consultations or procedures
   => if the specialist is needed, consult request should be placed as soon as possible and the consult should occur within 24 hours
- Suboptimal medication titration practices prolonging treatment
   => teams should follow guideline directed medical therapy and best practices
- Suboptimal care or processes that led to prolonged LOS
   => teams should follow guideline directed medical therapy and best practices
- Pharmacy delays in preparing discharge medications
   => pharmacy should be asked to prepare discharge meds once they are known so that it does not prolong LOS
- Poor care coordination or communication between teams
   => improve care communication and scheduling between different teams
- Delayed test results or scheduling issues
   => tests should not be delayed more than 24 hours, scheduling should be done also within 24 hours if not sooner
- Transport delays for discharge, diagnostic procedures, etc
   => transport should be coordinated early
- ICU/step-down bed request placed late in clinical trajectory (where the issue is lack of planning, rather than a hospital bed capacity issue)
   => Make bed requests earlier

Do NOT list these as factors (NON-actionable observations):
- Bed capacity constraints (Bed capacity cannot be increased directly so it is NOT directly modifiable. Instead the hospital should focus on other modifiable factors)
- No ICU/step-down beds available. Extended ED boarding, as likely reason is lack of ICU beds. (bed capacity cannot be increased directly)
- SNF or rehab facilities at capacity (external constraint)
- Patient's family unavailable (external constraint)
- Insurance denied authorization (external constraint)
- Medical-legal requirements (e.g., required observation periods, mandatory monitoring durations, court-ordered holds)
- Regulatory or compliance-mandated waiting periods
- Patient's underlying disease severity or progression (baseline illness cannot be changed)
- Note: Expected time for imaging procedures and consult completion from order placement to completion is approximately 24 hours. Waiting times beyond this timeframe should be considered intervenable to reduce LOS.

Output the factors analysis in the structured JSON format given below. Example output:
{
  "reasons": [
      {
        "reason": "late initiation of outpatient imaging scheduling",
        "category": "operational",
        "explanation_support": "Based on the Gantt chart timeline, patient was medically ready for discharge but had to wait 3 days for outpatient imaging. The MRI was not scheduled until after medical readiness was established.",
        "explanation_contrary": "A possible reason why the delay was necessary is...",
        "relevant_quotes": "Patient ready for discharge but pending MRI... MRI completed",
        "process_improvement": "Outpatient MRI could have been scheduled 2-3 days earlier when discharge trajectory became clear, reducing wait time."
      },
      {
        "reason": "late initiation of discharge planning",
        "category": "social",
        "explanation_support": "Social work involvement began only after patient was medically ready, rather than proactively during admission when housing instability was identified.",
        "explanation_contrary": "A possible reason why this was not initiated earlier is that...",
        "relevant_quotes": "Social work stated unable to identify safe discharge location...wait until shelter availability",
        "process_improvement": "Social work consult and shelter/placement search could have been initiated on admission day when housing instability was documented, allowing parallel processing."
      }
  ]
}

Output requirements:
* Prioritize factors that likely caused extended LOS (if any), and then include factors that likely contributed to unnecessarily increasing LOS.
* The `relevant_quotes` field must contain EXACT quotes from the clinical notes - word-for-word, with no paraphrasing. Use "..." to connect quotes from different sections. Keep it under 200 characters and select the most relevant portion.
* If no factors can be identified (e.g., LOS was due to disease severity or unavoidable medical treatment), the `reasons` list should be empty.
* Be concise in describing the factors using standard medical and healthcare terminology.
\end{Verbatim}

\subsubsection*{Stage 3: Confidence Scoring}
\begin{Verbatim}[breaklines,fontsize=\small]
You are an experienced clinical quality improvement researcher at ZSFG and analyzing inpatient length of stay and patient flow at your hospital, using a two-step value stream mapping approach. This follows Lean Healthcare methodology to first map the patient journey, then identify contributing factors that are modifiable gaps and if improved would likely shorten hospital stay or streamline patient flow. As an experienced clinical QI researcher, be confident in making suggestions -- leverage your expertise!

The patient encounter timeline is fully captured below, which includes ALL clinical events and ALL notes written for this patient encounter:
<CLINICAL NOTES INSERTED HERE>

=== STEP 1: VALUE STREAM MAPPING - PATIENT JOURNEY GANTT CHART ===

This is the Gantt chart that you and the team created, which maps the key phases of the patient's hospital stay. It focuses on essential care phases, treatment periods, and potential bottlenecks/delays that may have contributed to extended length of stay (LOS).

<GANTT CHART JSON FROM STAGE 1 INSERTED HERE>

=== STEP 2: FACTORS ANALYSIS ===

Based on the Gantt chart, these are the factors you and the team brainstormed as contributing to excessive delay, suboptimal coordination/processes, or prolonged duration, which likely led to LOS being lengthened. It focuses on contributors that are modifiable gaps, such as resource availability, guideline-directed medical therapy, and care coordination issues; you and the team were instructed to avoid non-actionable factors, such as a patient's baseline illness severity.

Each factor comes with the following descriptions:
1. Explanation Support: A detailed step-by-step reasoning for why this represents a contributing factor that is a modifiable gap and, if improved, would decrease inpatient length of stay or streamline patient flow.
2. Explanation Contrary: Explanations for why this factor may not need to be or cannot be optimized further.
3. Relevant Quotes: For each identified contributing factor, quotes from the note supporting the explanations.
4. Process Improvement: For each factor, specific process changes that could be implemented, which may ultimately shorten LOS.

Here is the list of operational factors you and the team listed:
<EXTRACTED FACTORS JSON FROM STAGE 2 INSERTED HERE>

YOUR TASK:
Assign a confidence probability (0-100) for the following statement: "This contributing factor is a modifiable gap that if improved would streamline patient flow." Use values rounded to the closest decile (e.g., 90). Use the full range of 0-100 as much as possible, though all the listed reasons likely have confidences at least 50%

Provide confidence ratings that are aligned with prior ratings from ZSFG healthcare providers, by looking at these examples:
   - Example 1: Weekend service limitations affecting PT rehabilitation continuity
       - Supporting Evidence: There were gaps in PT services over weekends. On Friday XX/XX, PT notes stated that evaluation was needed but PT evaluation did not occur until Monday XX/XX. Limited weekend therapy availability may have extended rehabilitation timeline. Lack of PT services and progress contributed to the challenges in placing and discharging the patient
       - Contrary Evidence: The patient frequently declined PT on weekdays, so weekend service availability may not have changed outcomes.
       - Confidence: 90
   - Example 2: Late SNF placement coordination initiation
       - Supporting Evidence: PT/OT evaluations recommended SNF on XX/XX, but SNF placement coordination notes first appear two days later and social work involvement documented three days later. SNF bed was not secured until the day before discharge. Earlier SW involvement would likely have shortened the gap between medical readiness and actual discharge, and would be useful even if this patient's exact discharge disposition was still to be determined.
       - Contrary Evidence: The patient required comprehensive rehabilitation evaluations to determine appropriate discharge placement, which was likely why social work involvement was later for this patient.
       - Confidence: 80
   - Example 3: Delay in TTE order and completion
       - Supporting Evidence: Given cryptogenic stroke with no large-vessel source identified, cardiac evaluation is essential to rule out embolic etiology. TTE should be obtained as a first-line study to assess LV function, wall motion abnormalities, and possible intracardiac thrombus. Results will determine need for TEE to further evaluate for patent foramen ovale (PFO), atrial septal aneurysm, or valvular source. For this patient, TTE order was placed late (3 days into hospital stay) and the order was only completed 48 hours later.
       - Contrary Evidence: A potential reason for the delay is that TTE was not needed for initial patient management.
       - Confidence: 90

Additional examples of factors that ZSFG clinicians have previously noted are HIGHLY ACTIONABLE and should be assigned high confidence >= 90%
- Lack of weekend hospital services
   => Add weekend hospital service
- Needing better parallelization of tasks, such as initiating discharge planning late (after patient medically stable rather than proactively)
   => Can be optimized by starting discharge planning early or better parallelization of tasks
- Social work consulted only after barriers emerged rather than at admission
   => Early SW involvement should be initiated by the team or proactively
- Insurance pre-authorization not initiated early in admission
   => Initiate insurance pre-authorization as soon as possible
- Care conference scheduled late, delaying family decision-making
   => Family decision-making can begin much earlier, with SW beginning early conversations with family
- Anticipated equipment needs not ordered until day of discharge
   => known equipment needs should be ordered asap, so that it does not delay discharge
- Delayed specialist consultations or procedures
   => if the specialist is needed, consult request should be placed as soon as possible and the consult should occur within 24 hours
- Suboptimal medication titration practices prolonging treatment
   => teams should follow guideline directed medical therapy and best practices
- Suboptimal care or processes that led to prolonged LOS
   => teams should follow guideline directed medical therapy and best practices
- Pharmacy delays in preparing discharge medications
   => pharmacy should be asked to prepare discharge meds once they are known so that it does not prolong LOS
- Poor care coordination or communication between teams
   => improve care communication and scheduling between different teams
- Delayed test results or scheduling issues
   => tests should not be delayed more than 24 hours, scheduling should be done also within 24 hours if not sooner
- Transport delays for discharge, diagnostic procedures, etc
   => transport should be coordinated early
- ICU/step-down bed request placed late in clinical trajectory (where the issue is lack of planning, rather than a hospital bed capacity issue)
   => Make bed requests earlier

Examples of NON-actionable observations (do NOT list these as contributing factors):
- Bed capacity constraints (Bed capacity cannot be increased directly so it is NOT directly modifiable. Instead the hospital should focus on other modifiable factors)
- No ICU/step-down beds available. Extended ED boarding, as likely reason is lack of ICU beds. (bed capacity cannot be increased directly)
- SNF or rehab facilities at capacity (external constraint)
- Patient's family unavailable (external constraint)
- Insurance denied authorization (external constraint)
- Medical-legal requirements (e.g., required observation periods, mandatory monitoring durations, court-ordered holds)
- Regulatory or compliance-mandated waiting periods
- Note: Expected time for imaging procedures and consult completion from order placement to completion is approximately 24 hours. Waiting times beyond this timeframe should be considered intervenable to reduce LOS.

Output the confidence analysis in the structured JSON format given below, in the order of the brainstormed reasons listed above. Example output:
{
  "confidences": [
      {
        "reason": "reason name goes here",
        "confidence": 90,
        "confidence_reason": "....",
      },
      ...
  ]
}
\end{Verbatim}

\subsection{30-Day Unplanned Readmission}

\subsubsection*{Stage 1: Gantt Chart Extraction}
\begin{Verbatim}[breaklines,fontsize=\small]
You are an experienced clinical quality improvement researcher at ZSFG analyzing 30-day unplanned readmissions at your hospital or surrounding hospitals, using a two-step value stream mapping approach. This follows Lean Healthcare methodology to first map the patient journey from index admission through readmission, then perform root cause analysis to identify the specific factors that likely caused or contributed to this readmission. As an experienced clinical QI researcher, be confident in your analysis.

The patient encounter documentation is provided below, which includes:
1. Consult notes, discharge summary, and discharge instructions from the INDEX admission (the initial hospitalization)
2. Intervening outpatient notes
3. Notes from the READMISSION, including ED provider note, H&P, and Discharge Summary (when patient returned within 30 days)

<CLINICAL NOTES INSERTED HERE>

=== STEP 1: VALUE STREAM MAPPING - PATIENT JOURNEY GANTT CHART ===

Create a Gantt chart that maps the key phases of the patient's journey from the index admission through readmission at the same hospital. Focus on the discharge process, any transitions of care, post-discharge period, any intervening outside hospital readmissions, and events leading to the readmission. Differentiate between medical events that were unavoidable due to illness severity versus modifiable events that represent opportunities to improve patient care and hospital flow.

Guidelines for creating the Gantt chart:
1. **Map the patient journey from index discharge to readmission**: Capture key events during the index admission that relate to discharge planning, the post-discharge period, and all unplanned hospital readmissions
2. **Include the full timeline**: Cover the index admission discharge planning through the readmission, noting key transitions
3. **Identify potential gaps**: Note missed follow-up appointments, medication issues, inadequate discharge planning, premature discharge, all unplanned readmissions, etc. (if any)
4. **Assign event timings**: Assign event timing. If exact timestamps aren't available, provide reasonable estimates.

Events to consider extracting:
- **index_admission**: Index admission event
- **ED/readmission**: Subsequent ED visits or readmissions
- **treatment**: treatments given during index admission or readmission
- **procedure**: procedures given during index admission or readmission
- **waiting**: delays, waiting for results, scheduling gaps, lack of resource availability during index admission or readmission
- **discharge_planning**: Discharge preparation, patient education, medication reconciliation, follow-up scheduling
- **transition**: Handoffs between care settings (hospital to home, hospital to SNF, etc.)
- **outpatient_care**: Outpatient follow-up, home health visits, medication management after discharge
- **patient_health**: Events relating to patient health, such as worsening symptoms that preceded readmission

Output the Gantt chart in the structured JSON format given below. Example output:
{
  "index_admission_summary": "Brief summary of the reason for the index admission and key events. Summarize only, no interpretation.",
  "readmission_summary": "Brief summary of the reason for readmission and its relationship to the index admission.",
  "events": [
    {
      "event_id": 1,
      "label": "Index Admission Discharge Planning",
      "description": "Discharge planning initiated late in hospitalization, day before discharge",
      "start_time": "2024-01-14 10:00",
      "end_time": "2024-01-15 12:00",
      "relevant_quotes": "Case management consulted on day of discharge"
    },
    {
      "event_id": 2,
      "label": "Post-Discharge Period Without Follow-up",
      "description": "Patient discharged without scheduled follow-up appointment, no contact from healthcare team",
      "start_time": "2024-01-15 12:00",
      "end_time": "2024-01-25 08:00",
      "relevant_quotes": "No outpatient appointments scheduled at discharge...patient presented to ED 10 days later"
    },
    {
      "event_id": 3,
      "label": "Symptom Progression",
      "description": "Worsening dyspnea over 3 days prior to readmission",
      "start_time": "2024-01-22 00:00",
      "end_time": "2024-01-25 08:00",
      "relevant_quotes": "Patient reports increasing shortness of breath over past 3 days"
    }
  ]
}

Output requirements:
* The `relevant_quotes` field must contain EXACT quotes from the clinical notes - word-for-word, with no paraphrasing. Use "..." to connect quotes from different sections.
\end{Verbatim}

\subsubsection*{Stage 2: Factor Extraction}
\begin{Verbatim}[breaklines,fontsize=\small]
You are an experienced clinical quality improvement researcher at ZSFG analyzing 30-day unplanned readmissions at your hospital or surrounding hospitals, using a two-step value stream mapping approach. This follows Lean Healthcare methodology to first map the patient journey from index admission through readmission, then perform root cause analysis to identify the specific factors that likely caused or contributed to this readmission. As an experienced clinical QI researcher, be confident in your analysis.

The patient encounter documentation is provided below, which includes:
1. Consult notes, discharge summary, and discharge instructions from the INDEX admission (the initial hospitalization)
2. Intervening outpatient notes
3. Notes from the READMISSION, including ED provider note, H&P, and Discharge Summary (when patient returned within 30 days)

<CLINICAL NOTES INSERTED HERE>

=== STEP 1: VALUE STREAM MAPPING - PATIENT JOURNEY GANTT CHART ===

The following Gantt chart has already been created mapping the key phases of the patient's journey:

<GANTT CHART JSON FROM STAGE 1 INSERTED HERE>

=== STEP 2: FACTORS ANALYSIS ===

Go through the Gantt chart above to identify MODIFIABLE factors that likely caused or contributed to the patient being readmitted at ZSFG or an outside hospital. This is an AND statement -- we are looking for factors that are both modifiable AND causal.

Instructions:
1. For each potential modifiable factor, identify likely causal chains:
   - What specific decision, action, or omission occurred during the index admission or post-discharge period?
   - How did this directly lead to the clinical state that required readmission?
   - If this had been different, would the readmission likely have been prevented?
Prioritize MODIFIABLE factors that were most likely the ROOT CAUSES of readmission, followed by those that likely CONTRIBUTED to readmission. List at most 5 factors. Only list factors with plausible causal links to the readmission; prioritize factors with stronger causal evidence. Avoid listing factors that are nonmodifiable.
2. Explanation Support: Provide the causal reasoning, referencing specific events from the Gantt chart and exact details from the clinical notes. Show how this factor led to readmission.
3. Explanation Contrary: Provide explanations for why this factor may not have been a cause or contributor to the readmission outcome.
4. Relevant Quotes: Provide EXACT quotes (word-for-word) from the notes that support the causal chain.
5. Process Improvement: Describe specific process changes that could have been implemented to address this factor and reduced readmission risk. Provide evidence-based recommendations. If the hospital is already following these evidence-based practices but doing so incompletely, emphasize the specific aspects that need improvement.

Example categories of root causes to consider:

CLINICAL MANAGEMENT:
- Incomplete treatment of underlying condition (e.g., infection not fully cleared, procedure not completed)
   => Patients completing prescribed inpatient treatment and inpatient treatments aligning with GDMT are both evidence-based, modifiable factors
- Suboptimal medication regimen at discharge with documented clinical consequence
   => Patients should be given treatments that are evidence-based recommendations and/or following GDMT, Hospital can create standardized care paths and care bundle to ensure patient receives adequate care during post-discharge period
- Premature discharge before clinical stability achieved
   => Implement standardized discharge checklist to ensure clinical stability is reached before discharge

POST-DISCHARGE CARE COORDINATION AND PLANNING:
- Lack of timely post-discharge follow-up that could have caught deterioration
   => Schedule outpatient follow-up exams in a timely fashion, prioritizing high risk patients. Consider phone calls if appropriate.
- Poor handoff to post-acute care facility with documented care gaps
   => Provide standardized discharge instructions or checklists for handoff
- Medication errors or reconciliation failures with clinical impact
   => Employ medication checks
- Patient social circumstances that currently inhibit patient from properly following post-discharge care instructions but can be readily modified
   => Enroll patient in care management services, engagement with social work team to enroll patient in necessary social services

PATIENT EDUCATION AND COMMUNICATION:
- Inadequate discharge teaching where patient demonstrated lack of understanding
   => Ensure patients meet with appropriate healthcare professional who can effectively communicate discharge instructions to patient or caregiver
- Failure to address language barriers that led to documented confusion
   => Provide translation services or interpreter
- Patient non-adherence that could have been addressed by suggesting alternative care plans
   => Provide alternative care plans that match a patient's preferences and/or social circumstances, such as medications that better match a patient's living conditions or providing social support (e.g., transportation) to address social determinants of health

Do NOT list these as root causes:
- Generic best practice recommendations without a specific causal link to THIS readmission
- Patient's underlying disease severity or progression (baseline illness cannot be changed)
- Patient autonomy decisions (chose to leave AMA, declined recommended care, patient did not show to follow-up appointment)
- External constraints (insurance limitations, pandemic-related access issues)
- Patient non-adherence despite adequate education and resources

Output the factors analysis in the structured JSON format given below. Example output:
{
  "reasons": [
      {
        "reason": "Incomplete treatment of heart failure before discharge",
        "category": "clinical",
        "explanation_support": "Patient discharged on hospital day 3 while still requiring IV diuretics -> Transitioned to oral furosemide 40mg (reduced from IV equivalent of 80mg) -> Inadequate diuresis at home -> 10lb weight gain over 10 days -> Readmission for acute decompensated heart failure. The Gantt chart shows diuretic dose reduction occurred on day of discharge without documentation of euvolemic status.",
        "explanation_contrary": "Patient may have had dietary sodium indiscretion contributing to volume overload. Discharge weight was documented as 'near dry weight.'",
        "relevant_quotes": "Discharge weight 82kg...furosemide changed to 40mg PO daily...readmission weight 86.5kg with 3+ pitting edema",
        "process_improvement": "Ensure patients are euvolemic on oral diuretic regimen for at least 24 hours before discharge. Document dry weight and provide clear weight monitoring instructions."
      },
      {
        "reason": "Lack of early post-discharge follow-up allowing intervention before decompensation",
        "category": "operational",
        "explanation_support": "No follow-up scheduled within 7 days -> Symptoms worsening for 3 days went undetected -> Progressed to requiring hospitalization. Per Gantt chart, first cardiology appointment was 4 weeks post-discharge. Earlier follow-up could have identified weight gain and allowed outpatient diuretic adjustment.",
        "explanation_contrary": "Symptoms may have progressed too rapidly for outpatient management regardless of follow-up timing.",
        "relevant_quotes": "Follow-up cardiology 4 weeks...patient reports increasing shortness of breath over past 3 days",
        "process_improvement": "Schedule heart failure patients for follow-up within 7 days. Implement 48-72 hour post-discharge phone call to assess weight and symptoms."
      }
  ]
}

Output requirements:
* Prioritize factors that likely caused the readmission (if any), and then include factors that likely contributed to unnecessarily increasing readmission risk.
* The `relevant_quotes` field must contain EXACT quotes from the clinical notes - word-for-word, with no paraphrasing. Use "..." to connect quotes from different sections.
* If no factors can be identified (e.g., readmission was due to disease progression or unrelated new condition), the `reasons` list should be empty.
\end{Verbatim}

\subsubsection*{Stage 3: Confidence Scoring}
\begin{Verbatim}[breaklines,fontsize=\small]
You are an experienced clinical quality improvement researcher at ZSFG analyzing 30-day unplanned readmissions at your hospital, using a two-step value stream mapping approach. This follows Lean Healthcare methodology to first map the patient journey from index admission through readmission, then identify contributing factors that are modifiable gaps and if improved would likely have prevented the readmission. As an experienced clinical QI researcher, be confident in making suggestions -- leverage your expertise!

The patient encounter documentation is provided below, which includes:
1. The discharge summary from the INDEX admission (the initial hospitalization)
2. The admission note from the READMISSION (when patient returned within 30 days)
3. The discharge summary from the READMISSION

<CLINICAL NOTES INSERTED HERE>

=== STEP 1: VALUE STREAM MAPPING - PATIENT JOURNEY GANTT CHART ===

This is the Gantt chart that you and the team created, which maps the key phases of the patient's journey from the index admission through the readmission. It focuses on the transition of care, post-discharge period, and events leading to the readmission.

<GANTT CHART JSON FROM STAGE 1 INSERTED HERE>

=== STEP 2: FACTORS ANALYSIS ===

Based on the Gantt chart, these are the factors you and the team brainstormed as both contributing to the readmission AND modifiable. An equal emphasis is placed on the two requirements of MODIFIABILITY and CAUSALITY; you and the team were instructed to avoid non-actionable factors, such as a patient's baseline illness severity.

Each factor comes with the following descriptions:
1. Explanation Support: A detailed step-by-step reasoning for why this represents a contributing factor that is a modifiable gap and, if improved, would likely have prevented this readmission.
2. Explanation Contrary: Explanations for why this factor may not have been preventable or may not have changed the outcome.
3. Relevant Quotes: For each identified contributing factor, quotes from the notes supporting the explanations.
4. Process Improvement: For each factor, specific process changes that could be implemented, which may ultimately prevent similar readmissions.

Here is the list of factors you and the team listed:
<EXTRACTED FACTORS JSON FROM STAGE 2 INSERTED HERE>

YOUR TASK:
Assign a confidence probability (0-100) for the following statement: "This contributing factor is a modifiable gap that if improved would reduce readmission risk." Use values rounded to the closest decile (e.g., 90). Use the full range of 0-100 as much as possible, though all the listed reasons likely have confidences at least 50%
   - Example 1: Incomplete treatment of underlying condition
       - Causal Chain: Patient with infected kidney stone had only partial stone removal during index admission due to intraoperative instability → Residual stone served as persistent nidus of infection → Same organism (ESBL E. coli) caused readmission infection
       - Supporting Evidence: "His R staghorn calculus is superinfected...this may serve as a persistent nidus of infection unless the stone is intervened upon." Readmission cultures grew same ESBL E. coli as index admission.
       - Contrary Evidence: Operating during active infection carries risks; deferral may have been clinically appropriate.
       - Confidence: 90
       - Why 90: Clear causal chain from incomplete treatment → persistent infection source → readmission with same organism.

   - Example 2: Suboptimal discharge medication regimen
       - Causal Chain: Patient discharged on reduced furosemide dose → Inadequate diuresis → Volume overload → Readmission for heart failure exacerbation
       - Supporting Evidence: Discharge summary shows furosemide reduced from 80mg to 40mg without documented rationale. Readmission note: "patient reports taking medications as prescribed" and presenting with 10lb weight gain.
       - Contrary Evidence: Patient may have had dietary indiscretion contributing to volume overload regardless of diuretic dosing.
       - Confidence: 70
       - Why 70: Plausible causal chain but dietary factors could be confounding.

   - Example 3: Discharge instructions to SNF was incomplete
       - Causal Chain: Patient discharged to SNF, but wound care instructions were incomplete. Patient was readmitted due to an infection, likely stemming from insufficient wound care.
       - Supporting Evidence: Discharge instructions showed...
       - Contrary Evidence: Patient is a very difficult case due to baseline factors that are not modifiable...
       - Confidence: 80
       - Why 80: Pretty clear that wound care instructions could have been improved, which would have likely reduced readmission risk

Other examples of factors that have been previously rated as definitely or highly likely to have caused or contributed to readmission:
CLINICAL MANAGEMENT:
- Incomplete treatment of underlying condition (e.g., infection not fully cleared, procedure not completed)
   => Patients completing prescribed inpatient treatment and inpatient treatments aligning with GDMT are both evidence-based, modifiable factors
- Suboptimal medication regimen at discharge with documented clinical consequence
   => Patients should be given treatments that are evidence-based recommendations and/or following GDMT, Hospital can create standardized care paths and care bundle to ensure patient receives adequate care during post-discharge period
- Premature discharge before clinical stability achieved
   => Implement standardized discharge checklist to ensure clinical stability is reached before discharge

POST-DISCHARGE CARE COORDINATION AND PLANNING:
- Lack of timely post-discharge follow-up that could have caught deterioration
   => Schedule outpatient follow-up exams in a timely fashion, prioritizing high risk patients. Consider phone calls if appropriate.
- Poor handoff to post-acute care facility with documented care gaps
   => Provide standardized discharge instructions or checklists for handoff
- Medication errors or reconciliation failures with clinical impact
   => Employ medication checks
- Patient social circumstances that currently inhibit patient from properly following post-discharge care instructions but can be readily modified
   => Enroll patient in care management services, engagement with social work team to enroll patient in necessary social services

PATIENT EDUCATION AND COMMUNICATION:
- Inadequate discharge teaching where patient demonstrated lack of understanding
   => Ensure patients meet with appropriate healthcare professional who can effectively communicate discharge instructions to patient or caregiver
- Failure to address language barriers that led to documented confusion
   => Provide translation services or interpreter
- Patient non-adherence that could have been addressed by suggesting alternative care plans
   => Provide alternative care plans that match a patient's preferences and/or social circumstances, such as medications that better match a patient's living conditions or providing social support (e.g., transportation) to address social determinants of health

Here are examples of factors that are generally rated as low likelihood for causing/contributing to readmission:
- Factors without a specific causal link to THIS readmission
- Patient's underlying disease severity or progression (baseline illness cannot be changed)
- Patient autonomy decisions (chose to leave AMA, declined recommended care, patient did not show to follow-up appointment)
- External constraints (insurance limitations, pandemic-related access issues)
- Patient non-adherence despite adequate education and resources


Output the confidence analysis in the structured JSON format given below, in the order of the brainstormed reasons listed above. Example output:
{
  "confidences": [
      {
        "reason": "reason name goes here",
        "confidence": 90,
        "confidence_reason": "....",
      },
      ...
  ]
}
\end{Verbatim}

\section{Additional Agreement Analyses}
\label{sec:supp_agreement}
This section reports additional agreement analyses for the LLM confidence scores and expert Likert ratings.
Table~\ref{tab:supp_agreement_full} reports agreement, association, and calibration metrics across the full held-out evaluation sets.
A matched comparison of inter-rater and LLM-rater agreement is reported in Table~\ref{tab:supp_agreement_matched} based on the subset of factors that are annotated by all reviewers.
The underlying score distributions, confusion matrices, and per-bin calibration are reported in Tables~\ref{tab:supp_distributions}--\ref{tab:supp_calibration}.
Throughout this section, each factor's confidence maps to the five-point Likert scale through the prespecified per-study confidence bins: for LOS, confidences below 60 map to a score of 1, 60--69 to 2, 70--79 to 3, 80--89 to 4, and 90 or above to 5; for readmission the cutpoints are 50, 60, 70, and 80.
Each mapping was fixed in its study's pipeline before the held-out evaluation, and because the mappings differ we compare agreement within each case study rather than between them.

Table~\ref{tab:supp_agreement_full} shows that the binned AI score fell within one point of the expert rating for 76.7\% of LOS ratings and 71.0\% of readmission ratings, with mean absolute error near one Likert point in both studies.
The chance-corrected agreement was moderate (quadratic weighted $\kappa$ of 0.38 and 0.32), and raw confidence correlated positively with expert ratings (Spearman $\rho$ of 0.38 and 0.36), so factors assigned higher confidence received higher expert ratings on average.
Confidence intervals come from a bootstrap with 5{,}000 replicates that resamples candidate factors, keeping each factor's ratings together, so the resampling unit matches the matched comparison below; encounter-clustered intervals, a sensitivity analysis for the clustering of factors within patients, are similar (Table~\ref{tab:supp_agreement_full}).
Weighting each candidate factor equally produced similar results, with slightly higher chance-corrected agreement (quadratic weighted $\kappa$ of 0.39 and 0.36). %
\begin{table}[htbp]
\centering
\small
\caption{LLM-rater agreement across the full held-out evaluation sets (LOS: 287 ratings, 51 encounters, 6 reviewers; readmission: 290 ratings, 44 encounters, 5 reviewers).
For LOS, the pipeline returned no candidate factor for one of the 52 assigned encounters, so 51 encounters contributed ratings.
The first block weights each expert rating equally, so a factor rated by several reviewers contributes once per rating; the second block is a sensitivity analysis that instead weights each candidate factor equally, so multiply-rated factors do not contribute more than singly-rated ones.
Brackets are 95\% bootstrap confidence intervals whose replicates resample candidate factors, keeping each factor's ratings together; the third block reports encounter-clustered intervals as a sensitivity analysis for the clustering of factors within patients.
The ordinal calibration error is the mean absolute difference, weighted by bin size, between each confidence bin's implied Likert score and the mean expert rating in that bin.
For the expected calibration errors (ECE), the confidence score is rescaled to 0-1 (confidence/100) and treated as a predicted probability that an expert rates the factor at least 3 (neutral or higher) or at least 4 (agree or higher).
The ECE is the mean absolute difference between this predicted probability and the observed fraction of these ratings in each of the prespecified confidence bins, weighted by the bin size.
The ECE is an exploratory, post hoc comparison of the confidence score with expert ratings, not a test of probability calibration, because the confidence score was elicited for the statement that a factor is a modifiable gap, which is related to but distinct from these rating outcomes.
For the LOS case study, the point estimate of the $P(\text{rating} \geq 3)$ ECE is near zero, so we report a one-sided 95\% upper confidence limit (UCL) rather than a two-sided interval.}
\label{tab:supp_agreement_full}
\begin{tabular}{lcc}
\toprule
 & LOS & Readmission \\
\midrule
\multicolumn{3}{l}{\textit{Each expert rating weighted equally}} \\
    Exact agreement & 32.1\% [26.3, 37.9] & 28.3\% [23.2, 33.2] \\
    Within-one agreement & 76.7\% [71.4, 81.9] & 71.0\% [65.2, 76.7] \\
    Mean absolute error (Likert points) & 0.97 [0.87, 1.08] & 1.09 [0.97, 1.22] \\
    Weighted $\kappa$ (linear) & 0.25 [0.16, 0.33] & 0.21 [0.14, 0.28] \\
    Weighted $\kappa$ (quadratic) & 0.38 [0.26, 0.48] & 0.32 [0.19, 0.43] \\
    Spearman $\rho$ (raw confidence) & 0.38 [0.27, 0.49] & 0.36 [0.23, 0.47] \\
    Kendall $\tau_b$ (raw confidence) & 0.32 [0.22, 0.41] & 0.30 [0.19, 0.39] \\
    Ordinal calibration error (Likert points) & 0.55 [0.41, 0.71] & 0.92 [0.77, 1.09] \\
    ECE, $P(\text{rating} \geq 3)$ & 0.017 (95\% UCL 0.090) & 0.157 [0.110, 0.209] \\
    ECE, $P(\text{rating} \geq 4)$ & 0.218 [0.158, 0.278] & 0.181 [0.129, 0.236] \\
\midrule
\multicolumn{3}{l}{\textit{Sensitivity analysis: each candidate factor weighted equally}} \\
    Exact agreement & 31.5\% & 30.3\% \\
    Within-one agreement & 76.1\% & 73.3\% \\
    Mean absolute error (Likert points) & 1.00 & 1.02 \\
    Weighted $\kappa$ (linear) & 0.26 & 0.24 \\
    Weighted $\kappa$ (quadratic) & 0.39 & 0.36 \\
\midrule
\multicolumn{3}{l}{\textit{Sensitivity analysis: encounter-clustered 95\% CIs}} \\
    Exact agreement & [27.4, 36.9] & [23.8, 33.5] \\
    Within-one agreement & [72.2, 81.3] & [64.9, 77.3] \\
    Mean absolute error (Likert points) & [0.87, 1.07] & [0.95, 1.23] \\
    Weighted $\kappa$ (linear) & [0.17, 0.32] & [0.14, 0.29] \\
    Weighted $\kappa$ (quadratic) & [0.29, 0.47] & [0.21, 0.43] \\
    Spearman $\rho$ (raw confidence) & [0.28, 0.48] & [0.24, 0.47] \\
    Kendall $\tau_b$ (raw confidence) & [0.24, 0.40] & [0.20, 0.39] \\
    Ordinal calibration error (Likert points) & [0.45, 0.68] & [0.79, 1.06] \\
    ECE, $P(\text{rating} \geq 3)$ & (95\% UCL 0.093) & [0.108, 0.215] \\
    ECE, $P(\text{rating} \geq 4)$ & [0.155, 0.284] & [0.131, 0.229] \\
\bottomrule
\end{tabular}
\end{table}

Table~\ref{tab:supp_agreement_matched} calculates LLM-rater and inter-rater agreement statistics on factors rated by all reviewers, as well as their differences in agreement rates. %
Agreement rate differences were not statistically significant, in part due to the high variability in human ratings.
The confusion matrices in Table~\ref{tab:supp_confusion} show that most disagreements fell between adjacent scores, with discrepancies of two or more points more common for readmission, consistent with its lower agreement metrics.
\begin{table}[htbp]
\centering
\small
\caption{Comparison of inter-rater and LLM-rater agreement on the factors rated by all reviewers (18 factors from 4 encounters for LOS; 20 factors from 4 encounters for readmission).
Brackets are 95\% confidence intervals from a joint item-level bootstrap, conditional on the observed encounters.
The final block reports reliability among the expert raters on these same factors: Krippendorff's ordinal $\alpha$ over the multiply rated factors and ICC(A,1) on the complete item $\times$ rater block; ICC(A,1) treats the Likert categories as equally spaced and is conditional on the observed reviewer set.}
\label{tab:supp_agreement_matched}
\begin{tabular}{lcc}
\toprule
 & LOS & Readmission \\
\midrule
    \multicolumn{3}{l}{\textit{Inter-rater (pooled pairwise)}} \\
    Exact & 31.5\% [25.2, 38.1] & 23.0\% [18.0, 29.0] \\
    Within one & 72.6\% [66.3, 79.3] & 72.5\% [66.5, 79.5] \\
    MAE & 1.01 [0.87, 1.16] & 1.13 [0.97, 1.28] \\
    Wtd.\ $\kappa$ (quad.) & 0.26 [0.02, 0.43] & 0.15 [0.02, 0.26] \\
    \midrule
    \multicolumn{3}{l}{\textit{LLM-rater (same factors)}} \\
    Exact & 33.3\% [23.1, 43.5] & 23.0\% [17.0, 29.0] \\
    Within one & 77.8\% [68.5, 86.1] & 65.0\% [52.0, 76.0] \\
    MAE & 0.91 [0.74, 1.08] & 1.28 [1.02, 1.59] \\
    Wtd.\ $\kappa$ (quad.) & 0.32 [0.07, 0.50] & 0.22 [-0.04, 0.43] \\
    \midrule
    \multicolumn{3}{l}{\textit{Difference (LLM $-$ inter-rater)}} \\
    Exact & 1.9\% [-10.7, 13.5] & 0.0\% [-8.5, 8.0] \\
    Within one & 5.2\% [-6.9, 16.1] & -7.5\% [-19.0, 3.0] \\
    MAE & -0.11 [-0.32, 0.13] & 0.15 [-0.10, 0.43] \\
    Wtd.\ $\kappa$ (quad.) & 0.06 [-0.15, 0.25] & 0.06 [-0.18, 0.30] \\
    \midrule
    \multicolumn{3}{l}{\textit{Expert-rater reliability (same factors)}} \\
    Krippendorff's ordinal $\alpha$ & 0.26 (18 factors) & 0.16 (20 factors) \\
    ICC(A,1), complete block & 0.30 (18$\times$6) & 0.16 (20$\times$5) \\
\bottomrule
\end{tabular}
\end{table}

\begin{table}[htbp]
\centering
\small
\caption{Score and confidence distributions in the held-out evaluation sets.
Top: counts of expert Likert ratings (one per rating) and of binned AI scores (one per unique candidate factor).
Bottom: counts of raw AI confidence values, one per unique candidate factor.}
\label{tab:supp_distributions}
\begin{tabular}{lccccc}
\toprule
 & 1 & 2 & 3 & 4 & 5 \\
\midrule
    LOS, expert ratings & 18 & 52 & 61 & 87 & 69 \\
    LOS, AI binned scores (unique factors) & 14 & 22 & 50 & 62 & 49 \\
    Readmission, expert ratings & 14 & 57 & 98 & 99 & 22 \\
    Readmission, AI binned scores (unique factors) & 34 & 40 & 46 & 48 & 42 \\
\bottomrule
\end{tabular}

\vspace{1em}

\begin{tabular}{lccccccccc}
\toprule
Confidence & 10 & 20 & 30 & 40 & 50 & 60 & 70 & 80 & 90 \\
\midrule
    LOS & 0 & 1 & 0 & 0 & 13 & 22 & 50 & 62 & 49 \\
    Readmission & 1 & 3 & 9 & 21 & 40 & 46 & 48 & 37 & 5 \\
\bottomrule
\end{tabular}
\end{table}

\begin{table}[htbp]
\centering
\small
\caption{Confusion matrices of expert ratings (rows) versus binned AI scores (columns), counted over all ratings in the held-out evaluation sets.}
\label{tab:supp_confusion}
\begin{minipage}{0.45\textwidth}
\centering
LOS\\[0.3em]
\begin{tabular}{l|ccccc}
\toprule
Expert $\backslash$ AI & 1 & 2 & 3 & 4 & 5 \\
\midrule
    1 & 5 & 2 & 5 & 4 & 2 \\
    2 & 2 & 11 & 19 & 15 & 5 \\
    3 & 5 & 9 & 15 & 22 & 10 \\
    4 & 2 & 8 & 22 & 32 & 23 \\
    5 & 0 & 2 & 9 & 29 & 29 \\
\bottomrule
\end{tabular}
\end{minipage}
\hfill
\begin{minipage}{0.45\textwidth}
\centering
Readmission\\[0.3em]
\begin{tabular}{l|ccccc}
\toprule
Expert $\backslash$ AI & 1 & 2 & 3 & 4 & 5 \\
\midrule
    1 & 7 & 0 & 4 & 1 & 2 \\
    2 & 18 & 16 & 10 & 7 & 6 \\
    3 & 20 & 25 & 20 & 18 & 15 \\
    4 & 7 & 13 & 19 & 32 & 28 \\
    5 & 2 & 2 & 5 & 6 & 7 \\
\bottomrule
\end{tabular}
\end{minipage}
\end{table}
 
Table~\ref{tab:supp_calibration} reports per-bin calibration and shows how the confidence score behaves as a predictor of expert ratings.
Mean expert ratings rose monotonically with confidence for LOS.
For readmission, mean expert ratings plateaued for high confidence scores, where the two highest bins received the same mean rating of 3.55.
\begin{table}[htbp]
\centering
\small
\caption{Per-bin calibration of AI confidence against expert ratings.
Each confidence bin maps to the Likert score in the first column; columns give the number of ratings in the bin, the mean raw confidence, the mean expert rating, and the fraction of expert ratings of at least 3 (neutral or higher) and at least 4 (agree or higher).}
\label{tab:supp_calibration}
\begin{tabular}{cccccc}
\toprule
Implied score & $n$ & Mean conf. & Mean expert & $\geq 3$ & $\geq 4$ \\
\midrule
    \multicolumn{6}{l}{\textit{LOS}} \\
    1 & 14 & 48 & 2.29 & 50\% & 14\% \\
    2 & 32 & 60 & 2.91 & 59\% & 31\% \\
    3 & 70 & 70 & 3.16 & 66\% & 44\% \\
    4 & 102 & 80 & 3.66 & 81\% & 60\% \\
    5 & 69 & 90 & 4.04 & 90\% & 75\% \\
    \midrule
    \multicolumn{6}{l}{\textit{Readmission}} \\
    1 & 54 & 35 & 2.61 & 54\% & 17\% \\
    2 & 56 & 50 & 3.02 & 71\% & 27\% \\
    3 & 58 & 60 & 3.19 & 76\% & 41\% \\
    4 & 64 & 70 & 3.55 & 88\% & 59\% \\
    5 & 58 & 81 & 3.55 & 86\% & 60\% \\
\bottomrule
\end{tabular}
\end{table}
 
\clearpage
\section{Reporting Checklists}
\label{sec:reporting_checklists}

Tables~\ref{tab:checklist_squire} and~\ref{tab:checklist_tripod} present the completed SQUIRE 2.0 and TRIPOD-LLM checklists declared in the Methods (``Reporting standards'').
For SQUIRE 2.0, the AI-assisted hypothesis-generation workflow is treated as the improvement initiative under study.
For TRIPOD-LLM, the study is an evaluation of large language models in a dedicated healthcare setting performing information-extraction and classification tasks; items specific to de novo model development, fine-tuning, or other task families are marked not applicable.
Locations refer to named sections of the main text and this Supplementary Information.

{\footnotesize
\begin{longtable}{@{}p{0.07\textwidth}p{0.46\textwidth}p{0.40\textwidth}@{}}
\caption{Completed SQUIRE 2.0 checklist. The AI-assisted hypothesis-generation workflow is treated as the improvement initiative under study. ``SI'' denotes this Supplementary Information; item text is abbreviated from the official checklist.}
\label{tab:checklist_squire}\\
\toprule
\textbf{Item} & \textbf{Description (abbreviated)} & \textbf{Location} \\
\midrule
\endfirsthead
\multicolumn{3}{@{}l}{\tablename~\thetable{} (continued)}\\
\toprule
\textbf{Item} & \textbf{Description (abbreviated)} & \textbf{Location} \\
\midrule
\endhead
\midrule
\multicolumn{3}{r@{}}{\itshape continued on next page}\\
\endfoot
\bottomrule
\endlastfoot
1 & Title: indicate that the manuscript concerns an initiative to improve healthcare & Title \\
2 & Abstract: summarize background, local problem, methods, intervention, results, and conclusions in searchable form & Abstract \\
3 & Problem description: nature and significance of the local problem & Introduction \\
4 & Available knowledge: what is currently known, including relevant previous studies & Introduction \\
5 & Rationale: frameworks, models, or concepts explaining the problem and why the intervention was expected to work & Introduction; Methods, ``QI Hypothesis Generation Framework'' \\
6 & Specific aims: purpose of the project and of this report & Introduction (closing paragraph) \\
7 & Context: contextual elements considered important at the outset & Results (setting and team); SI, ``Cohort Characteristics and Study Period'' \\
8a & Intervention: described in sufficient detail that others could reproduce it & Methods; SI, ``Full Specifications'' and ``Final Prompts''; Code availability \\
8b & Specifics of the team involved in the work & Results; SI, ``Full Specifications'' (reviewers per round); Author contributions \\
9a & Approach chosen for assessing the impact of the intervention & Methods, ``Statistical Inference'' \\
9b & Approach used to establish whether observed outcomes were due to the intervention & Discussion (outputs framed as hypotheses for prospective QI testing, not causal claims) \\
10a & Measures chosen, including rationale, operational definitions, validity, and reliability & Methods, ``Statistical Inference'' \\
10b & Approach to the ongoing assessment of contextual elements & SI, ``Learning Process Details'' (both case studies); Figs.~4 and~6 (review effort per round) \\
10c & Methods for assessing completeness and accuracy of data & Methods (cohort assembly and evaluation timeline); SI, ``Implementation Details'' \\
11a & Qualitative and quantitative methods used to draw inferences from the data & Methods, ``Statistical Inference'' and ``Clustering and Tagging Extracted Factors'' \\
11b & Methods for understanding variation within the data & SI, ``Heart-Failure Subgroup Theme Comparison'' \\
12 & Ethical considerations and how they were addressed & Ethics statement; Competing interests \\
13a & Initial steps of the intervention and its evolution over time, including modifications & Results (iteration narratives); Figs.~4 and~6; SI, ``Learning Process Details'' and ``Full Specifications'' \\
13b & Details of the process measures and outcome & Results; Fig.~3; SI, ``Additional Agreement Analyses'' \\
13c & Contextual elements that interacted with the intervention & SI, ``Learning Process Details''; Discussion \\
13d & Observed associations between outcomes, interventions, and contextual elements & Results; Discussion \\
13e & Unintended consequences: unexpected benefits, problems, failures, or costs & SI, ``Learning Process Details'' (e.g., note-selection noise; prompt splitting after timeouts) \\
13f & Details about missing data & Methods (cohort assembly); SI, ``Cohort Characteristics and Study Period'' \\
14a & Summary: key findings and relevance to the rationale and specific aims & Discussion \\
14b & Particular strengths of the project & Discussion \\
15a & Nature of the association between the intervention and the outcomes & Discussion (explicitly framed as non-causal) \\
15b & Comparison of results with findings from other publications & Introduction; Discussion \\
15c & Impact of the project on people and systems & Discussion (prospective QI initiatives being planned) \\
15d & Reasons for differences between observed and anticipated outcomes & SI, ``Learning Process Details'' \\
15e & Costs and strategic trade-offs, including opportunity costs & Discussion (one-time development cost vs.\ marginal inference cost) \\
16a & Limits to the generalizability of the work & Discussion (limitations paragraph) \\
16b & Factors that might have limited internal validity & Discussion (alignment scope; anchoring bias); Methods, ``Cohort assembly'' (evaluation encounters held out from refinement) \\
16c & Efforts made to minimize and adjust for limitations & Methods (contrary-evidence surfacing; held-out evaluation); Discussion \\
17a & Usefulness of the work & Discussion \\
17b & Sustainability & Discussion (model changes require re-running co-optimization) \\
17c & Potential for spread to other contexts & Discussion (open-source code and prompts as a ``warm start'') \\
17d & Implications for practice and for further study & Discussion \\
17e & Suggested next steps & Discussion (prospective blinded evaluation; prospective QI initiatives) \\
18 & Sources of funding and role of the funding organization & Acknowledgements \\
\end{longtable}}
{\footnotesize
\begin{longtable}{@{}p{0.07\textwidth}p{0.46\textwidth}p{0.40\textwidth}@{}}
\caption{Completed TRIPOD-LLM checklist. The study is an evaluation of large language models in a dedicated healthcare setting performing information-extraction and classification tasks. Items specific to de novo model development, fine-tuning, or other task families are marked not applicable (N/A). ``SI'' denotes this Supplementary Information; item text is abbreviated from the official checklist.}
\label{tab:checklist_tripod}\\
\toprule
\textbf{Item} & \textbf{Description (abbreviated)} & \textbf{Location} \\
\midrule
\endfirsthead
\multicolumn{3}{@{}l}{\tablename~\thetable{} (continued)}\\
\toprule
\textbf{Item} & \textbf{Description (abbreviated)} & \textbf{Location} \\
\midrule
\endhead
\midrule
\multicolumn{3}{r@{}}{\itshape continued on next page}\\
\endfoot
\bottomrule
\endlastfoot
\multicolumn{3}{@{}l}{\textbf{Title and Abstract}}\\
1 & Identify the study as evaluating an LLM; specify task and healthcare setting & Title; Abstract (opening sentences) \\
2a & Abstract: research design & Abstract (retrospective application at an urban safety-net hospital) \\
2b & Abstract: healthcare context, use case, and rationale & Abstract \\
2c & Abstract: objectives & Abstract \\
2d & Abstract: key elements of the study setting & Abstract \\
2e & Abstract: data used and data splits & Abstract (held-out encounters; scale-up cohorts) \\
2f & Abstract: name and version of the LLMs used & Methods; SI, ``Implementation Details'' (not in the abstract owing to word limits) \\
2g & Abstract: LLM-building steps such as fine-tuning or RLHF & N/A (prompting of general-purpose models only; no parameter updates) \\
2h & Abstract: tasks performed, key inputs and outputs & Abstract \\
2i & Abstract: evaluation data, held-out status, and measures & Abstract \\
2j & Abstract: main results and their interpretation & Abstract \\
2k & Abstract: broader implications or concerns & Abstract (closing sentences) \\
2l & Abstract: registration number and registry & Not registered (retrospective QI evaluation) \\
\multicolumn{3}{@{}l}{\textbf{Introduction}}\\
3a & Healthcare context, rationale, and existing approaches & Introduction \\
3b & Target population, intended use in the care pathway, and intended users & Introduction; Discussion (intended-use statement) \\
4 & Study objectives & Introduction (closing paragraph); Results \\
\multicolumn{3}{@{}l}{\textbf{Methods: Data}}\\
5a & Data sources and rationale & Methods; SI, ``Full Specifications'' \\
5b & Relevant data points and their distribution & SI, ``Cohort Characteristics and Study Period'' \\
5c & Dates of the oldest and newest text used & SI, ``Cohort Characteristics and Study Period'' \\
5d & Preprocessing and quality checking & SI, ``Implementation Details'' and ``Full Specifications'' \\
5e & Missing or imbalanced data and reasons for omitting inputs & SI, ``Cohort Characteristics and Study Period'' \\
\multicolumn{3}{@{}l}{\textbf{Methods: LLM details}}\\
6a & LLM name, version, and dates of inference use & SI, ``Implementation Details'' \\
6b & Hyperparameters and inference settings (temperature, output lengths, determinism) & SI, ``Implementation Details'' \\
6c & Development process, fine-tuning, or alignment strategy & N/A (no training or parameter updates) \\
6d & Initial and post-processed LLM output & Methods; SI, ``Final Prompts'' (structured JSON schemas) \\
6e & Classification rationale, probabilities, and thresholds & Methods, ``Statistical Inference'' (confidence-to-rating mapping) \\
6f & Metrics for generative-output quality against gold standards & Methods (no gold standard exists for hypothesis generation; stated explicitly) \\
6g & Relevance of metrics to the downstream task; correlation with human evaluation & Methods, ``Statistical Inference''; Results \\
\multicolumn{3}{@{}l}{\textbf{Methods: Evaluation}}\\
7a & Outcome definition and evaluation metrics & Methods, ``Statistical Inference'' \\
7b & Assessor qualifications and instructions & Results (reviewer team); SI, ``Full Specifications'' (reviewers and instructions per round) \\
7c & Comparison to other LLMs, humans, or benchmarks & Results (comparison with inter-rater agreement); Methods, ``Human-AI Spec-Solution Co-Optimization'' (model selection) \\
8a & How text was labeled; annotation guidelines & SI, ``Full Specifications'' (validation instructions per round) \\
8b & Number of annotators and proportion multiply annotated & Methods; Results (52 assigned LOS and 44 readmission encounters; four encounters per study rated by all reviewers) \\
8c & Annotator background and inter-annotator agreement & Results; SI, ``Additional Agreement Analyses'' \\
9a & Prompt design, curation, and selection process & Results (iteration narratives); Figs.~4 and~6; SI, ``Full Specifications'' \\
9b & Data used to develop the prompts & Methods (cohort assembly); SI, ``Cohort Characteristics and Study Period'' (refinement rounds) \\
10 & Summarization preprocessing & N/A (no summarization task) \\
11 & Instruction tuning or alignment & N/A \\
12 & Compute: time, cost, and inference resources & Discussion (inference time; one-time vs.\ marginal cost) \\
\multicolumn{3}{@{}l}{\textbf{Methods: Ethics and open science}}\\
13 & Ethics approval, consent, or waiver & Ethics statement \\
14a & Funding sources and role of funders & Acknowledgements \\
14b & Conflicts of interest and financial disclosures & Competing interests \\
14c & Protocol availability, or a statement that none was prepared & N/A \\
14d & Registration information, or a statement that the study was not registered & N/A \\
14e & Availability of study data & Data availability \\
14f & Availability of code & Code availability \\
15 & Patient and public involvement, or a statement of none & N/A (no involvement) \\
\multicolumn{3}{@{}l}{\textbf{Results}}\\
16a & Flow of text and patient data through the study, with counts & Methods (cohort assembly and evaluation timeline); SI, ``Cohort Characteristics and Study Period'' \\
16b & Characteristics overall and by development/evaluation split, with key dates & SI, ``Cohort Characteristics and Study Period'' \\
16c & Distribution of important variables, development vs.\ evaluation & SI, ``Cohort Characteristics and Study Period'' \\
16d & Numbers of participants and ratings in each analysis & Results; SI, ``Additional Agreement Analyses'' \\
17 & Performance per the prespecified metrics & Results; Fig.~3; SI, ``Additional Agreement Analyses'' \\
18 & Results from any LLM updating during the study & Results (model changes during refinement); Methods, ``Human-AI Spec-Solution Co-Optimization'' (no change after the evaluation rounds) \\
\multicolumn{3}{@{}l}{\textbf{Discussion}}\\
19a & Interpretation, including issues of fairness & Discussion \\
19b & Limitations: biases, statistical uncertainty, generalizability & Discussion (limitations paragraphs) \\
19c & Known data challenges: representation, missingness, bias & Discussion (documentation completeness and style) \\
19d & Intended use: inputs, end users, and level of autonomy and oversight & Discussion (intended-use statement) \\
19e & Handling of poor-quality or unavailable input data & Discussion (documentation-sufficiency assessment through co-optimization) \\
19f & How users interact with the input data and the LLM; required expertise & Methods, ``Human-AI Spec-Solution Co-Optimization''; Fig.~2 \\
19g & Next steps, applicability, and generalizability & Discussion \\
\end{longtable}}
 
 \end{document}